\documentclass[11pt, a4paper]{thuc3i}
\usepackage[sort&compress]{natbib}

\setheadertext{Recursive Harness Self-Improvement}
\makeatletter
\def\@selectedlogo{0}
\makeatother

\usepackage{hyperref}
\usepackage{url}
\usepackage{tabularx}
\usepackage{makecell}
\usepackage{array,ragged2e,booktabs}
\usepackage{amsfonts}
\usepackage{amsthm}
\usepackage{nicefrac}
\usepackage{microtype}
\usepackage{xcolor}
\usepackage{algorithm}
\usepackage{algpseudocode}

\definecolor{darkblue}{rgb}{0, 0, 0.5}
\hypersetup{colorlinks=true, citecolor=darkblue, linkcolor=darkblue, urlcolor=darkblue}

\usepackage{xurl}
\usepackage{amsmath}
\usepackage{graphicx}
\usepackage{mathtools}
\usepackage{pifont}
\usepackage{colortbl}
\usepackage{xspace}
\usepackage{multirow}
\usepackage{enumitem}
\usepackage{caption}
\usepackage{float}

\renewcommand{\today}{2026-10-01}
\usepackage[dvipsnames]{xcolor}
\definecolor{paperInk}{HTML}{263C50}
\definecolor{paperMuted}{HTML}{677C8E}
\definecolor{paperLink}{HTML}{376A8E}
\definecolor{paperBlue}{HTML}{A6C7DD}
\definecolor{paperBlueLight}{HTML}{F1F8FB}
\definecolor{paperPurple}{HTML}{8981C4}
\definecolor{paperPurpleDark}{HTML}{7367B2}
\definecolor{paperPurpleLight}{HTML}{F0EDFA}
\definecolor{paperPurpleSoft}{HTML}{D8D2EE}
\definecolor{paperGrid}{HTML}{E5EEF4}
\definecolor{paperBaseline}{HTML}{EEF3F6}
\definecolor{paperStatic}{HTML}{E8F5FA}
\definecolor{paperRecursive}{HTML}{E8EEF9}
\definecolor{paperGain}{HTML}{E7F4EF}
\definecolor{paperLoss}{HTML}{F8EDEF}
\definecolor{paperChartBlue}{HTML}{5888C6}
\definecolor{paperChartBlueLight}{HTML}{C6D9F0}
\definecolor{paperChartCyan}{HTML}{3B9CB5}
\definecolor{paperChartCyanLight}{HTML}{BDE3EA}
\definecolor{paperChartMint}{HTML}{5AA384}
\definecolor{paperChartMintLight}{HTML}{CBE7D9}

\title{Recursive Harness Self-Improvement for Frontier Reasoning Data Synthesis}

\author{%
    Wenlong Zhang$^{*1,2}$,
    Zhengbo Jiao$^{*1,3}$,
    Chenxu Zhang$^{*4}$,
    Lekang Jiang$^{5}$,
    SiYuan Ma$^{6}$,
    Qituan Zhang$^{1,7}$,
    Guo Chen$^{1,8}$,
    Linfeng Zhang$^{\dagger1}$\\
    $^{1}$EPIC Lab, Shanghai Jiao Tong University \quad
    $^{2}$University of Science and Technology of China \quad
    $^{3}$Shanghai University of Finance and Economics \quad
    $^{4}$Imperial College London \quad
    $^{5}$University of Cambridge \quad
    $^{6}$Nanyang Technological University \quad
    $^{7}$Fudan University \quad
    $^{8}$University of Electronic Science and Technology of China
    \vskip -1mm
    \textbf{$^*$Equal Contribution.} \quad \textbf{$^\dagger$Corresponding Author.}
}

\begin{abstract}
Generating progressively harder reasoning problems requires synthesis
procedures that adapt as the task distribution evolves. Existing
task-level recursion reuses generated problems as seeds but leaves the
construction harness unchanged. We present task--harness co-evolution,
a framework for recursive harness self-improvement (RSI) in reasoning-data
synthesis. Online self-improvement converts intermediate solver failures
into reusable skills during generation. Post-task self-improvement
revises skills, prompts, and workflows after each batch, adopting
candidates only when they generate harder valid tasks within a bounded
cost increase. Model weights and verification criteria remain fixed.
Across mathematics, coding, and science, mean solver accuracy decreases
from 100.0\% to 54.8\% over fourteen evolution rounds. Ablations show
that combining both update schedules produces harder tasks than
fixed-harness recursion or either schedule alone. The resulting data
improves downstream SFT and GRPO performance. In particular, a 27B
student fine-tuned on 10K synthesized mathematics examples achieves
62.5\% mean-16 accuracy on APEX, competitive with selected frontier-model
references. These results support adapting the synthesis harness
alongside the tasks to generate increasingly challenging data with
downstream training value.
\end{abstract}

\begin{document}

\maketitle

\begin{figure}[t]
    \centering
    \includegraphics[width=\linewidth]{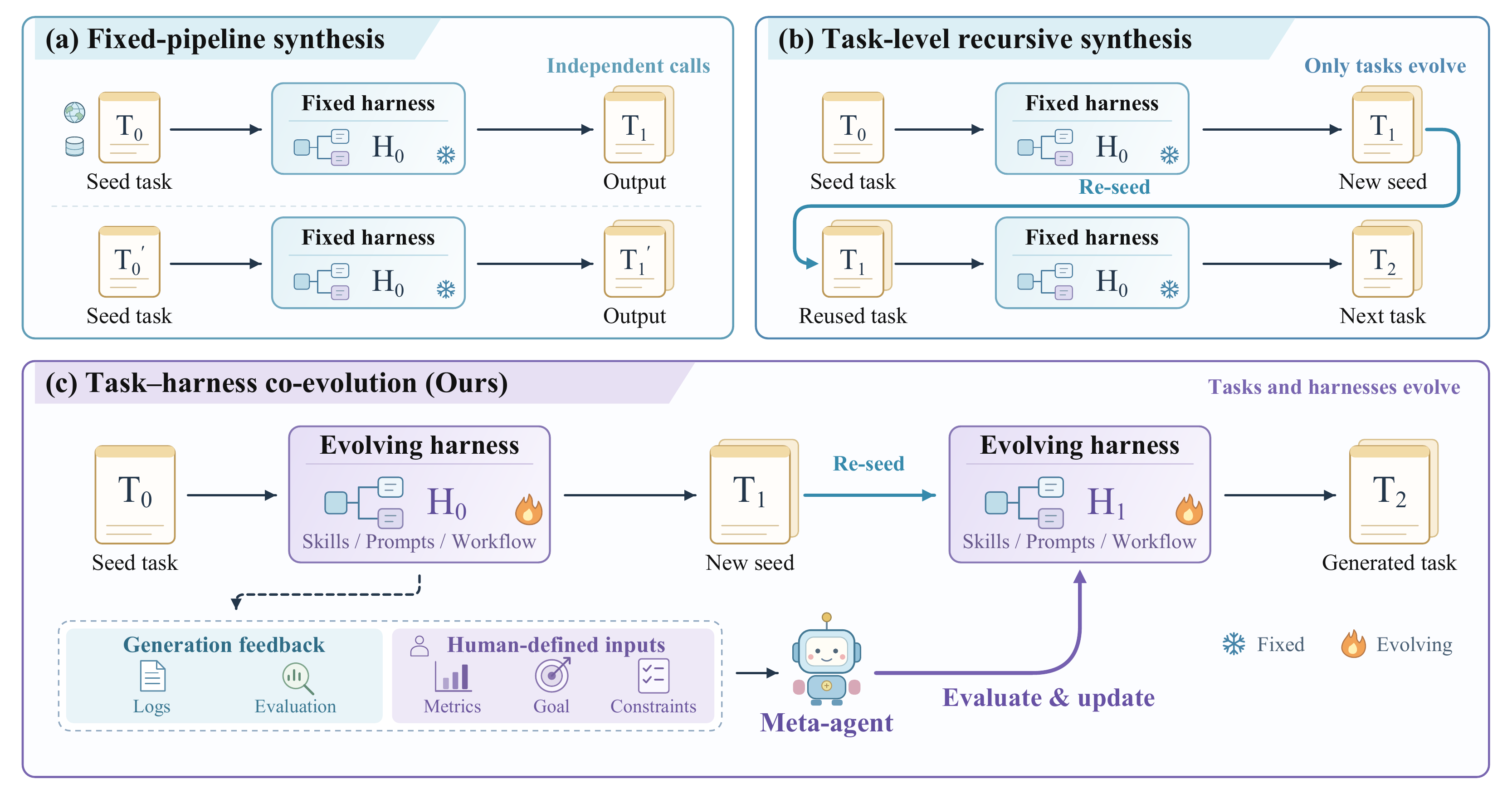}
    \caption{\textbf{Three reasoning-data synthesis regimes.}
    \textbf{(a)} Fixed-pipeline synthesis maps independent seeds to outputs
    through the same harness $H_0$.
    \textbf{(b)} Task-level recursion feeds each generated task back as
    the next seed, while construction skills and workflows stay fixed.
    \textbf{(c)} Our approach recursively updates both tasks and the harness.
    A meta-agent uses generation logs and evaluation feedback, together
    with human-defined goals and update constraints, to revise skills,
    prompts, and workflows. Evaluated revisions guide subsequent generation
    under fixed verification criteria. Snowflakes and flames denote fixed
    and evolving harnesses, respectively.}
    \label{fig:synthesis-paradigms}
\end{figure}

\section{Introduction}
\label{sec:introduction}

Advances in mathematical, scientific, and coding reasoning depend on
a continuing supply of high-quality, challenging problems.
OpenMathInstruct-2 demonstrates the value of synthetic data for
mathematical reasoning training~\citep{toshniwal2025openmathinstruct2},
while FrontierMath highlights the need for problems that distinguish
advanced reasoning capabilities~\citep{glazer2024frontiermath}.
As model capabilities improve, however, a static dataset cannot adapt
its difficulty. Increasing dataset size alone does not ensure that the
synthesis procedure can construct progressively harder reasoning problems.
Here, the synthesis harness comprises reusable skills, prompts, and
workflows that turn a seed into a verified problem. Adapting these
components offers a way to respond to changing solver weaknesses.

Existing synthesis methods progress from instruction expansion to
recursive task evolution. Self-Instruct generates instructions from
model outputs~\citep{wang2023selfinstruct}, and WizardLM increases
instruction complexity through predefined operators~\citep{xu2024wizardlm}.
RST recursively reuses verified terminal tasks as seeds, demonstrating
the potential of task-level recursion under a largely fixed synthesis
procedure~\citep{li2026rst}. Yet evolving the tasks alone leaves their
reusable construction skills and workflows unchanged. Autodata and
DataFoundry instead optimize data-generating agents and data-preparation
runtimes~\citep{kulikov2026autodata,yang2026datafoundry}. Together, these
directions motivate adapting the generation procedure alongside its
evolving task distribution.
This raises a specific question for recursive synthesis: can the
generation procedure improve as its own outputs become the inputs
to subsequent rounds?

To address this limitation, we present \textbf{task--harness co-evolution},
a framework that improves reasoning tasks and their synthesis harness
together (Figure~\ref{fig:synthesis-paradigms}). Verified tasks seed the
next round, while generation feedback guides two complementary update
schedules. Online self-improvement converts intermediate solver failures
into reusable skills before generation finishes. Post-task self-improvement
uses completed-batch feedback to revise skills, prompts, and workflows.
Candidate harnesses are compared with the online snapshot before adoption,
with rollback when revisions do not satisfy the validity, difficulty,
and cost criteria. Model weights and verification logic remain fixed,
so adaptation occurs in the generation procedure rather than the model.
Candidate evaluation uses the harness containing online updates as
its baseline. If all batch-level candidates are rejected, online skills
are retained for the next generation round.

\begin{figure*}[!t]
    \centering
    \includegraphics[width=\linewidth]{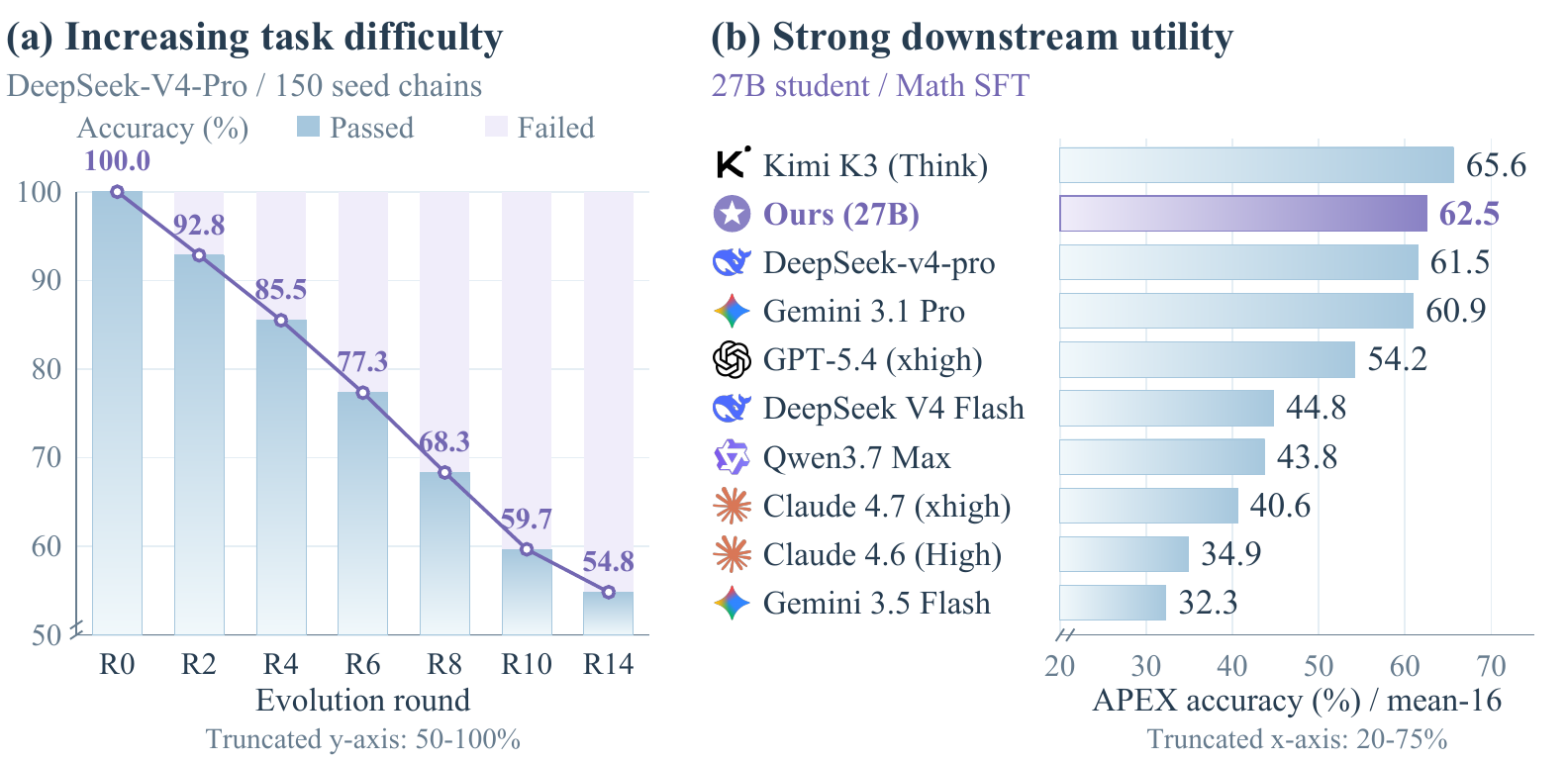}
    \caption{\textbf{Harder tasks, useful training data.}
    \textbf{(a)} DeepSeek-V4-Pro accuracy falls from 100.0\% to 54.8\%
    across 150 seed chains (50 per domain, equally weighted).
    The line tracks Passed; Failed is its complement; the y-axis is truncated at 50\%.
    \textbf{(b)} SFT on 10K synthesized math examples yields \textbf{62.5\%}
    APEX accuracy for a 27B student, competitive with selected frontier models.
    We evaluate all APEX models with 16 attempts per problem; the x-axis starts at 20\%.
    \mbox{APEX evaluation details: Appendix~\ref{sec:apex-reference}.}}
    \label{fig:intro-results}
\end{figure*}

Experiments across mathematics, coding, and science demonstrate sustained
difficulty growth and downstream training value
(Figure~\ref{fig:intro-results}). Fixed-solver accuracy falls from
100.0\% to 54.8\% over fourteen evolution rounds, while the synthesized
data improves both SFT and GRPO performance. With only 10K mathematics
examples, a 27B student reaches 62.5\% on APEX, competitive with selected
frontier-model references on this benchmark. Our main contributions
are as follows:
\begin{enumerate}
    \item \textbf{Task--Harness Co-Evolution.} We introduce a synthesis
    framework that recursively evolves reasoning tasks and their
    generation harness together, extending task-only recursion to
    adaptive construction procedures.

    \item \textbf{Online and Post-Task Self-Improvement.} We combine
    immediate skill updates during generation with batch-level harness
    revision, candidate evaluation, and rollback under fixed verification
    criteria and bounded cost growth.

    \item \textbf{Difficulty Growth and Training Utility.} We show
    that the combined update schedules produce harder tasks than
    fixed-harness recursion and either schedule alone, while the resulting
    data improves downstream reasoning through both SFT and GRPO.
\end{enumerate}

\section{Related Work}

\paragraph{Frontier Reasoning in Science.}
The progress of large language models has made the difficulty frontier
of reasoning a moving target. Mathematical and scientific reasoning
benchmarks such as MATH, GPQA, and FrontierMath progressively test competition-level,
graduate-level, and exceptionally difficult reasoning
\citep{hendrycks2021math,rein2024gpqa,glazer2024frontiermath}.
Humanity's Last Exam and SciCode extend this evaluation to broad
academic knowledge and research-oriented scientific coding
\citep{phan2025humanity,tian2024scicode}, while FrontierScience
explicitly evaluates expert-level scientific tasks
\citep{wang2026frontierscience}. Complementary solver systems such as
AlphaGeometry, AlphaProof, DeepSeek-R1, and AlphaEvolve combine
language models with formal verification, reinforcement learning, or
automated search to tackle increasingly complex mathematical and
scientific problems
\citep{trinh2024alphageometry,hubert2026alphaproof,
guo2025deepseekr1,novikov2025alphaevolve}. Together, these benchmarks
and systems show that progress requires problems that expose the
remaining capabilities of strong solvers. They also motivate synthesis
methods that can generate valid and informative challenges as the
reasoning frontier moves.

\paragraph{Reasoning Data Synthesis.}
Synthetic-data methods have evolved from instruction bootstrapping to
feedback-driven and recursive task construction. Self-Instruct samples
and filters model-generated instructions, while WizardLM applies
predefined operators to increase instruction complexity
\citep{wang2023selfinstruct,xu2024wizardlm}. OpenMathInstruct-2 further
studies large-scale math question--solution synthesis and its effect on
downstream training \citep{toshniwal2025openmathinstruct2}. Other
methods improve reasoning data through self-training or self-play:
STaR filters generated rationales by correctness, ReST$^{\mathrm{EM}}$
iterates between generation and model updating, and Absolute Zero and
R-Zero use verifiable feedback to couple task proposal with reasoning
improvement \citep{zelikman2022star,singh2024restem,
zhao2025absolutezero,huang2025rzero}. Auto Evol-Instruct learns
instruction-evolution strategies from prior trajectories
\citep{zeng2024autoevol}. BenchEvolver and Recursive Synthetic Terminal
Tasks (RST) make recursion explicit by reusing solutions or verified
task bundles as seeds for subsequent task generation
\citep{wu2026benchevolver,li2026rst}. This line of work shows that
task or seed recursion can expand the difficulty frontier. Our approach
further adapts the construction skills and workflow alongside the
evolving task distribution.

\paragraph{Harness Evolution and RSI.}
Another line of work treats an agent's surrounding program as an
optimization target. ADAS generates and evaluates code-defined agent
designs, while AFlow searches over code-represented workflows using
execution feedback \citep{hu2024adas,zhang2024aflow}. Recursive
self-improvement methods further evolve the improvement procedure
itself: Promptbreeder evolves mutation prompts, STOP recursively
improves a code optimizer, and the Darwin G\"odel Machine maintains an
archive of validated self-modifications
\citep{fernando2023promptbreeder,zelikman2023stop,zhang2025dgm}.
Recent harness-oriented systems study prompt-level revision, end-to-end
harness optimization, and meta-agents that improve both a task agent
and its improvement procedure
\citep{lee2026rhi,lee2026metaharness,zhang2026hyperagents}.
Autodata optimizes a data-scientist scaffold using generation and solver
feedback, whereas DataFoundry evolves an executable data-preparation
runtime through pilot-set diagnosis and
component-level adapters \citep{kulikov2026autodata,yang2026datafoundry}.

\section{Harness-Level Self-Evolution}
\label{sec:method}

\begin{figure*}[!t]
    \centering
    \includegraphics[width=\linewidth]{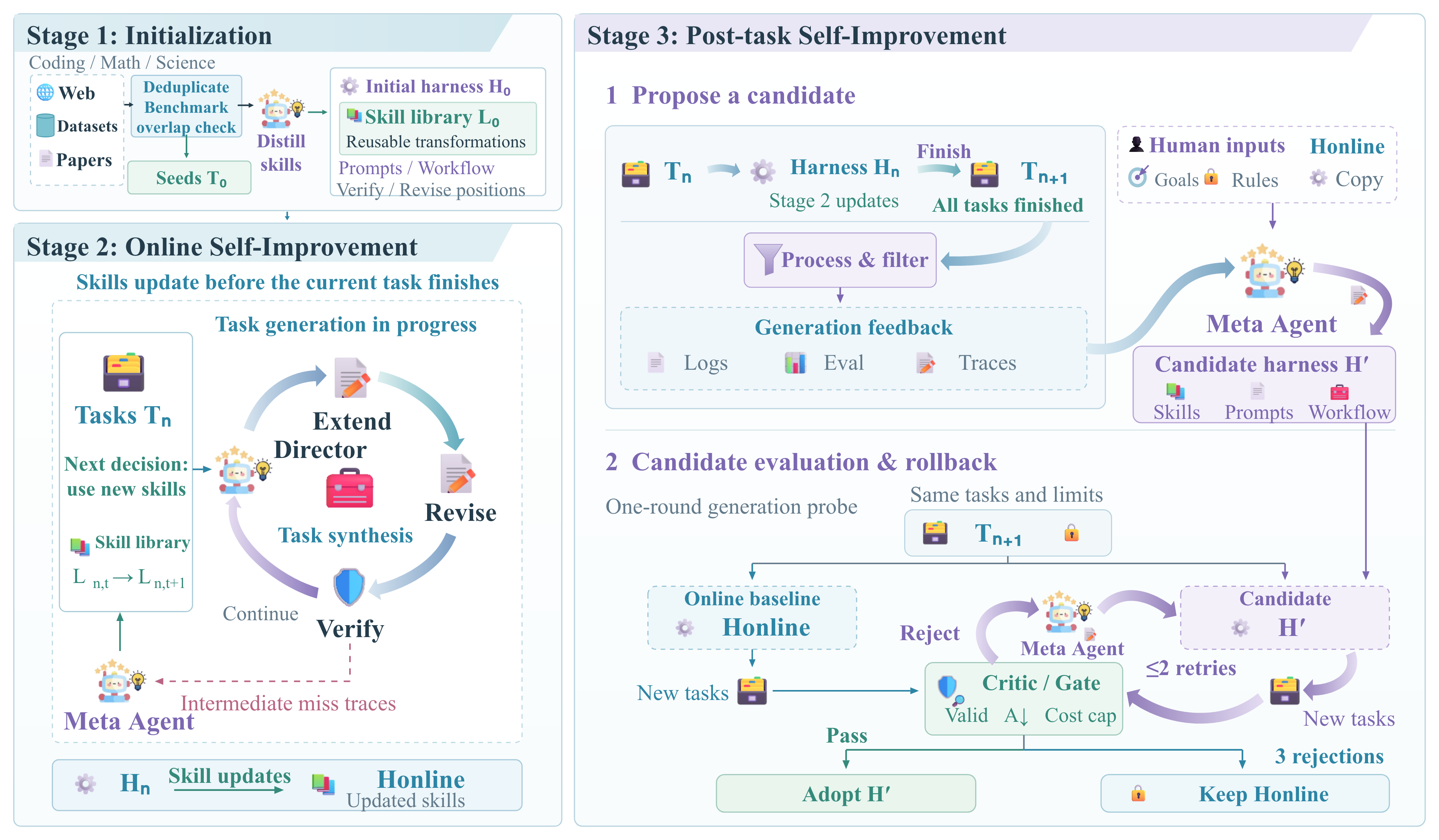}
    \caption{\textbf{Three-stage task--harness co-evolution.}
    \textbf{Stage~1} initializes screened seeds and $H_0$.
    \textbf{Stage~2} updates skills during generation, yielding
    $H_n^{\mathrm{online}}$.
    \textbf{Stage~3} processes feedback after batch completion or budget
    exhaustion and uses human inputs to propose a candidate (top).
    Candidate and online baseline each synthesize one round from shared
    current tasks under matched limits (bottom). Adopt the first passing
    candidate; three rejections (initial proposal plus two retries) retain
    the online snapshot and discard post-task edits. The gate requires
    lower solver accuracy with bounded cost growth. Model weights and
    verifier logic remain fixed.}
    \label{fig:harness}
\end{figure*}

We recursively synthesize tasks with a fixed language model and an
adaptive harness (Figure~\ref{fig:harness}). After one-time initialization
(Stage~1), each round combines two update schedules. Online
self-improvement (Stage~2) turns intermediate solver failures into
immediately usable construction skills during generation. Post-task
self-improvement (Stage~3) revises, evaluates, and selects the harness
after the batch ends. Algorithm~\ref{alg:evolution}
(Appendix~\ref{sec:method-algorithm}) gives the complete procedure.

\subsection{Problem Formulation and Initialization}
\label{sec:formulation}

A harness comprises reusable construction skills $\mathcal L$, role
prompts $P$, workflow $W$, and verification/revision placements
$\rho_V,\rho_R$:
\begin{equation}
H=(\mathcal L,P,W,\rho_V,\rho_R).
\label{eq:harness}
\end{equation}
Model weights, verifier implementation, and final validity criteria
remain fixed. Online adaptation changes the skill component; batch-level
optimization can revise all components, including when verification
and revision are invoked.

Starting from $N$ verified seeds $\mathcal T_0$, round $n$ constructs
$\mathcal T_{n+1}$ from $\mathcal T_n$ using a working copy of $H_n$.
Each seed chain retains its latest valid, completed QA. An unsuccessful
attempt preserves the predecessor, so every round contains one task
per original seed. We do not select the hardest ancestor retrospectively;
failed attempts remain in the logs and cost accounting.

For a matched cohort, let $q_i(H)$ be chain $i$'s retained task and
$\widehat p(q)\in[0,1]$ its fraction of correct answers over four attempts
under a fixed solver configuration. Define mean accuracy, generation
cost per seed, and relative difficulty against the online baseline $H_o$ as
\begin{equation}
\begin{aligned}
\bar A(H)&=\frac{1}{N}\sum_{i=1}^{N}\widehat p(q_i(H)),\\
\bar C_{\mathrm{gen}}(H)&=\frac{C_{\mathrm{gen}}(H)}{N},\\
r_D(H',H_o)&=\frac{\bar A(H_o)}{\bar A(H')}.
\end{aligned}
\label{eq:evaluation-metrics}
\end{equation}
$r_D>1$ indicates harder tasks; it is a ratio of cohort means, not a mean
of per-task ratios. $C_{\mathrm{gen}}$ includes Director, generation,
verification, revision, and online-update calls, including failed attempts.
We prioritize difficulty subject to validity and bounded cost growth,
allowing additional computation for harder tasks. Repeated solver attempts
are not independent evolution runs.

\paragraph{Stage 1: data curation and initialization.}
Before selecting seeds or distilling skills, we screen candidate seeds
and source passages from coding, mathematics, and science datasets,
papers, and web documents for duplicates and benchmark overlap.
Duplicate seed families remain within one evolution/held-out split;
benchmark items and solutions are isolated from the generation agents.
An acquisition model extracts reusable task transformations using the
fields \texttt{intent}, \texttt{method},
\texttt{difficulty\_effect}, and \texttt{tool\_hint}. Screened skills,
initial prompts, and workflow form $H_0$. Skill examples and training
exports are checked again for overlap, since input screening alone does
not exclude contamination introduced during synthesis
\citep{toshniwal2025openmathinstruct2}. Appendix~\ref{sec:method-protocol}
details curation and evaluation procedures.

\subsection{Online self-improvement}
\label{sec:online}

The Director selects skills and controls \textsc{Extend}, \textsc{Revise},
and \textsc{Verify} until \textsc{Finish} or budget exhaustion.
The diagram shows the initial ordering, which batch-level optimization
may revise. During verification, the generating model also attempts
the intermediate QA $q_{n,t}$. A solver failure attributable to a
reasoning weakness on a valid QA yields a trace $z^{\mathrm{miss}}_{n,t}$.
The Meta Agent converts this weakness into a reusable construction
transformation, rather than an answer to that particular QA:
\begin{equation}
\mathcal L_{n,t+1}=\operatorname{Admit}\!\left(
\mathcal L_{n,t},
\operatorname{Meta}_{s}(q_{n,t},z^{\mathrm{miss}}_{n,t};\Gamma_s)
\right).
\label{eq:online}
\end{equation}
$\Gamma_s$ specifies applicability, reusability, and supporting evidence.
\textsc{Admit} incorporates qualifying skills; invalid QA, incorrect
reference answers, and tool failures instead trigger repair.

The next Director call reads the updated library, making the skill
available to the unfinished task and remaining tasks in the same batch.
This avoids delaying reusable failure feedback until all tasks finish.
At the batch boundary, we freeze the resulting harness as
$H_n^{\mathrm{online}}$, the comparison baseline and fallback for the
next optimization step. Admitted skills need not increase task difficulty.

\subsection{Post-task self-improvement}
\label{sec:batch}

Post-task self-improvement starts after the task batch completes or its generation
budget expires. We treat revision and candidate evaluation as a single outer optimization
loop, using completed batches to assess effects beyond an individual
failure trace. This organization follows feedback-driven harness search
\citep{lee2026metaharness,lee2026rhi}; we evaluate task validity,
difficulty, and generation cost.

\paragraph{Feedback and revision.}
Stage~3's production row recaps Stage~2 rather than generating again.
After batch termination, in-loop tests of retained tasks
$\mathcal T_{n+1}$ produce scores $\mathcal E_n$ and failure traces.
Process \& filter prepares these signals and run logs for the
Meta Agent without excluding failed attempts from evaluation or costs.
The Meta Agent first updates skills in a copy of
$H_n^{\mathrm{online}}$, yielding $\widehat H_n$. It then uses execution
logs $\mathcal Z_n$, operational metrics $\mathcal M_n$, and human-specified
goal $g$ and constraints $\Gamma$ to propose a candidate:
\begin{equation}
\begin{aligned}
\Delta H_n^{(j)}&=\operatorname{Meta}_{h}
(\widehat H_n,\mathcal Z_n,\mathcal M_n,\mathcal E_n;
\Gamma,g,\mathcal R_{<j}),\\
H_n^{\prime(j)}&=\operatorname{Apply}
(\widehat H_n,\Delta H_n^{(j)}).
\end{aligned}
\label{eq:proposal}
\end{equation}
The goal and constraints are specified in advance. Allowed edits include
adding, revising, or removing skills; changing prompts and workflow;
and repositioning \textsc{Verify} and \textsc{Revise}. Verifier logic
cannot change. Each proposal starts from $\widehat H_n$ and can use
previous rejection reports $\mathcal R_{<j}$; rejected edits do not
accumulate automatically.

\paragraph{Candidate evaluation and selection.}
We freeze a shared cohort $S_n$ of current retained tasks as evaluation
inputs. Each candidate and $H_o=H_n^{\mathrm{online}}$ independently
synthesize one further round from $S_n$, with identical model/tool
settings and resource ceilings. We compare the new
outputs, not previously recorded scores of $S_n$. Each side uses its
own library and online adaptation in isolation; evaluation updates and
outputs do not alter the production snapshots or task chains.
Let $Q(H)$ be the fraction of new final candidates passing validity
before fallback, with missing candidates counted as invalid. The Critic
accepts a candidate only if
\begin{equation}
\begin{aligned}
A(H',H_o)=\mathbb I\bigl[&H'\in\mathcal H_\Gamma,
\ Q(H')\ge Q(H_o),\\
&\bar A(H')<\bar A(H_o),\\
&\bar C_{\mathrm{gen}}(H')\le
\min\{1+1.5(r_D-1),\,1.5\}\\
&\qquad {}\cdot \bar C_{\mathrm{gen}}(H_o)\bigr],
\end{aligned}
\label{eq:acceptance}
\end{equation}
where $\mathcal H_\Gamma$ enforces edit and fixed resource limits.
The prescribed cost allowance is 1.5 times the relative difficulty gain,
capped at a 50\% increase over the online baseline. Thus a 20\% difficulty
gain permits at most 30\% higher cost. This difficulty-first empirical
gate requires no validity decrease; it guarantees neither improved
cost efficiency nor improvement on unseen tasks. Appendix~\ref{sec:method-protocol}
specifies cohort composition, evaluation budgets, boundary cases, and accounting.

The first accepted candidate becomes $H_{n+1}$. Input tasks and baseline
results stay fixed across the initial proposal and at most two retries
(three candidate evaluations). After three rejections,
$H_{n+1}=H_n^{\mathrm{online}}$: all post-task edits, including
post-batch skill updates, are discarded. Online skills and rejection
logs remain. Exhausting the evaluation budget without acceptance also
retains the online snapshot; incomplete probes cannot pass. We archive
accepted and online-only fallback versions with
distinct statuses; both use $\mathcal T_{n+1}$ to seed the next round.
Final held-out evaluations never inform updates or candidate selection.

\section{Experiments}
\label{sec:experiments}

\begin{table*}[!t]
    \centering
        \centering
    \caption{\textbf{GRPO with Qwen3.5-4B Thinking.} Mean-16 accuracy (\%; higher
    is better). Overall averages the four benchmarks; superscripts give
    percentage-point changes from the checkpoint. Changes are computed
    from the displayed scores. Bold marks column-best scores.}
    \label{tab:grpo-4b}
    \begingroup
    \small
    \colorlet{grpoBaseline}{paperBaseline}
    \colorlet{grpoStatic}{paperStatic}
    \colorlet{grpoRecursive}{paperRecursive}
    \colorlet{grpoHarness}{paperPurpleLight}
    \colorlet{grpoOverall}{paperBlueLight}
    \colorlet{grpoGain}{paperGain}
    \colorlet{grpoLoss}{paperLoss}
    \newcommand{\grponumber}[2]{%
        \makebox[20pt][r]{#1}\makebox[21pt][l]{#2}}
    \newcommand{\grpobadge}[4]{%
        \grponumber{#1}{\raisebox{1.8pt}{\begingroup
        \setlength{\fboxsep}{0.5pt}%
        \colorbox{#2}{\fontsize{7}{8}\selectfont
        \ensuremath{\textstyle #3}\normalfont\fontsize{7}{8}\selectfont #4}%
        \endgroup}}}
    \newcommand{\grpoup}[2]{\grpobadge{#1}{grpoGain}{\uparrow}{#2}}
    \newcommand{\grpodown}[2]{\grpobadge{#1}{grpoLoss}{\downarrow}{#2}}
    \newcommand{\grpooverall}[1]{\begingroup
        \setlength{\fboxsep}{0.6pt}\colorbox{grpoOverall}{#1}\endgroup}
    \newcommand{\grpogroup}[2]{%
        \multicolumn{6}{@{}l@{}}{\begingroup
        \setlength{\fboxsep}{2pt}%
        \makebox[0pt][l]{\colorbox{#1}{\parbox[c]{\dimexpr\linewidth-4pt\relax}{\bfseries #2}}}%
        \endgroup}\\[1pt]}
    \setlength{\tabcolsep}{2pt}
    \renewcommand{\arraystretch}{1.02}
    \begin{tabular*}{\linewidth}{@{\extracolsep{\fill}}lcccc|c@{}}
        \toprule
        \textbf{Training data} & \textbf{APEX Shortlist}
        & \textbf{HMMT Feb 2025}
        & \textbf{LCB v6} & \textbf{SciBench} & \textbf{Overall} \\
        \midrule
        \grpogroup{grpoBaseline}{Qwen3.5-4B Thinking (Baseline Model)}
        No additional training & \grponumber{32.4}{} & \grponumber{73.5}{}
        & \grponumber{55.8}{} & \grponumber{71.8}{} & \grpooverall{\grponumber{58.4}{}} \\
        \addlinespace[3pt]
        \grpogroup{grpoStatic}{Static Data Synthesis}
        OpenMathInstruct-2 & \grpoup{33.6}{1.2} & \grpoup{75.4}{1.9}
        & \grpodown{52.6}{3.2} & \grpodown{70.9}{0.9} & \grpooverall{\grpodown{58.1}{0.3}} \\
        OpenCodeReasoning & \grpodown{27.5}{4.9} & \grpodown{70.8}{2.7}
        & \grpoup{58.7}{2.9} & \grpodown{71.6}{0.2} & \grpooverall{\grpodown{57.2}{1.2}} \\
        WildSci & \grpodown{28.2}{4.2} & \grpodown{70.0}{3.5}
        & \grpodown{53.6}{2.2} & \grpodown{70.6}{1.2} & \grpooverall{\grpodown{55.6}{2.8}} \\
        MetaMath & \grpodown{29.3}{3.1} & \grpodown{68.1}{5.4}
        & \grpodown{52.2}{3.6} & \grpodown{68.4}{3.4} & \grpooverall{\grpodown{54.5}{3.9}} \\
        \addlinespace[3pt]
        \grpogroup{grpoRecursive}{Task-Level Recursive Synthesis}
        IQC & \grpoup{34.4}{2.0} & \grpoup{75.0}{1.5}
        & \grpodown{55.2}{0.6} & \grpoup{73.5}{1.7} & \grpooverall{\grpoup{59.5}{1.1}} \\
        \addlinespace[3pt]
        \grpogroup{grpoHarness}{Harness-Level Self-Evolution}
        \textbf{Ours} & \grpoup{\textbf{36.4}}{4.0} & \grpoup{\textbf{78.3}}{4.8}
        & \grpoup{\textbf{59.2}}{3.4} & \grpoup{\textbf{76.9}}{5.1}
        & \grpooverall{\grpoup{\textbf{62.7}}{4.3}} \\
        \bottomrule
    \end{tabular*}
    \endgroup

    \par\vspace{8pt}
        \centering
    \caption{\textbf{SFT with Qwen3.5-35B-A3B.} Mean-16 accuracy
    (\%; higher is better), using 12K examples per data configuration.
    Ours uses 10K mathematics and 2K coding examples. Overall is the
    unweighted mean of the four benchmark columns. Superscripts give
    percentage-point changes from the checkpoint, computed from
    the displayed scores. Bold marks column-best scores.}
    \label{tab:sft-35b}
    \begingroup
    \small
    \newcommand{\sftnumber}[2]{%
        \makebox[20pt][r]{#1}\makebox[21pt][l]{#2}}
    \newcommand{\sftbadge}[4]{%
        \sftnumber{#1}{\raisebox{1.8pt}{\begingroup
        \setlength{\fboxsep}{0.5pt}%
        \colorbox{#2}{\fontsize{7}{8}\selectfont
        \ensuremath{\textstyle #3}\normalfont\fontsize{7}{8}\selectfont #4}%
        \endgroup}}}
    \newcommand{\sftup}[2]{\sftbadge{#1}{paperGain}{\uparrow}{#2}}
    \newcommand{\sftdown}[2]{\sftbadge{#1}{paperLoss}{\downarrow}{#2}}
    \newcommand{\sftflat}[2]{\sftbadge{#1}{paperBaseline}{=}{#2}}
    \newcommand{\sftoverall}[1]{\begingroup
        \setlength{\fboxsep}{0.6pt}\colorbox{paperBlueLight}{#1}\endgroup}
    \newcommand{\sftgroup}[2]{%
        \multicolumn{6}{@{}l@{}}{\begingroup
        \setlength{\fboxsep}{2pt}%
        \makebox[0pt][l]{\colorbox{#1}{\parbox[c]{\dimexpr\linewidth-4pt\relax}{\bfseries #2}}}%
        \endgroup}\\[1pt]}
    \setlength{\tabcolsep}{2pt}
    \renewcommand{\arraystretch}{1.02}
    \begin{tabular*}{\linewidth}{@{\extracolsep{\fill}}lcccc|c@{}}
        \toprule
        \textbf{Training data} & \textbf{APEX} & \textbf{APEX Shortlist}
        & \textbf{HMMT Feb 2026} & \textbf{LCB v6} & \textbf{Overall} \\
        \midrule
        \sftgroup{paperBaseline}{Qwen3.5-35B-A3B (Baseline Model)}
        No additional training & \sftnumber{4.2}{} & \sftnumber{44.3}{}
        & \sftnumber{80.9}{} & \sftnumber{74.8}{}
        & \sftoverall{\sftnumber{51.1}{}} \\
        \addlinespace[3pt]
        \sftgroup{paperGain}{Baselines}
        + OpenMathInstruct-2 & \sftdown{3.6}{0.6} & \sftup{45.3}{1.0}
        & \sftup{81.8}{0.9} & \sftdown{72.2}{2.6}
        & \sftoverall{\sftdown{50.7}{0.4}} \\
        + OpenCodeReasoning & \sftdown{2.6}{1.6} & \sftdown{40.4}{3.9}
        & \sftdown{75.6}{5.3} & \sftup{77.5}{2.7}
        & \sftoverall{\sftdown{49.0}{2.1}} \\
        + MMIQC / IQC & \sftdown{3.1}{1.1} & \sftup{46.3}{2.0}
        & \sftup{82.2}{1.3} & \sftdown{72.9}{1.9}
        & \sftoverall{\sftflat{51.1}{0.0}} \\
        \addlinespace[3pt]
        \sftgroup{paperPurpleLight}{Harness-Level Self-Evolution (Ours)}
        \textbf{+ Ours} & \sftup{\textbf{7.3}}{3.1} & \sftup{\textbf{49.7}}{5.4}
        & \sftup{\textbf{86.7}}{5.8} & \sftup{\textbf{79.1}}{4.3}
        & \sftoverall{\sftup{\textbf{55.7}}{4.6}} \\
        \bottomrule
    \end{tabular*}
    \endgroup

\end{table*}

\subsection{Experimental Setup}
\label{sec:setup}

We evaluate synthesis across mathematics, coding, and science using
DeepSeek-V4-Pro as the fixed solver; difficulty is measured by mean
accuracy over four attempts per task. Downstream experiments train
Qwen3.5-4B Thinking with GRPO and Qwen3.5-35B-A3B and Qwen3.8-27B
with SFT. We evaluate the benchmarks in Tables~\ref{tab:grpo-4b}
and~\ref{tab:sft-35b} using MathArena~\citep{dekoninck2026matharena}
and report mean-16 accuracy. Within each training comparison,
baselines share hyperparameters. Model assignments, training
configurations, and evaluation details are consolidated in
Appendix~\ref{sec:experimental-setup}.

\subsection{Difficulty Progression}
\label{sec:difficulty}

We evolve 150 seed chains for 14 rounds, with 50 per domain. Round
$r=0$ denotes the original seeds; unsuccessful attempts retain the
latest valid task. Overall accuracy equally weights domains; repeated
solver attempts are not independent evolution runs.

\begin{figure*}[!t]
    \centering
    \includegraphics[width=\linewidth]{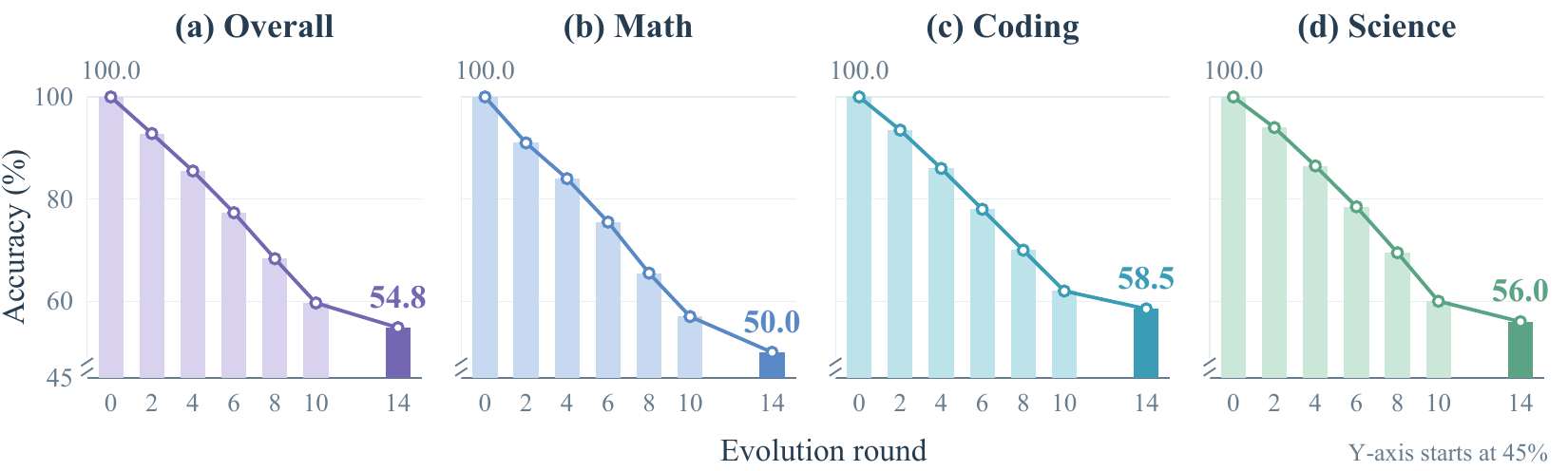}
    \caption{\textbf{Difficulty progression.} DeepSeek-V4-Pro accuracy
    (\%; lower is harder) over (a) all 150 seed chains and (b--d) three
    domains of 50 seeds each. The shared y-axis spans 45--100\%; full
    values are in Appendix~\ref{sec:evolution-records}.}
    \label{fig:domain-progression}
\end{figure*}

Overall accuracy decreases from 100.0\% to 54.8\% by round~14,
a 45.2-point decline. Domain accuracies reach 50.0\%, 58.5\%, and
56.0\%, respectively (Figure~\ref{fig:domain-progression}). The decline
occurs in every domain under a fixed cohort, rather than through a
change in domain composition.

\subsection{Online and Post-task Self-Improvement Ablations}
\label{sec:ablations}

Four configurations share 50 initial seeds and harness: Task-only
keeps the harness fixed; Online only updates skills during
generation; Post-task only updates the harness after each batch,
including skill revisions and candidate evaluation; Full
combines both schedules (Sections~\ref{sec:online}--\ref{sec:batch}). All
retain QA verification, independently of the harness acceptance gate.

\begin{figure*}[!t]
    \centering
    \includegraphics[width=\linewidth]{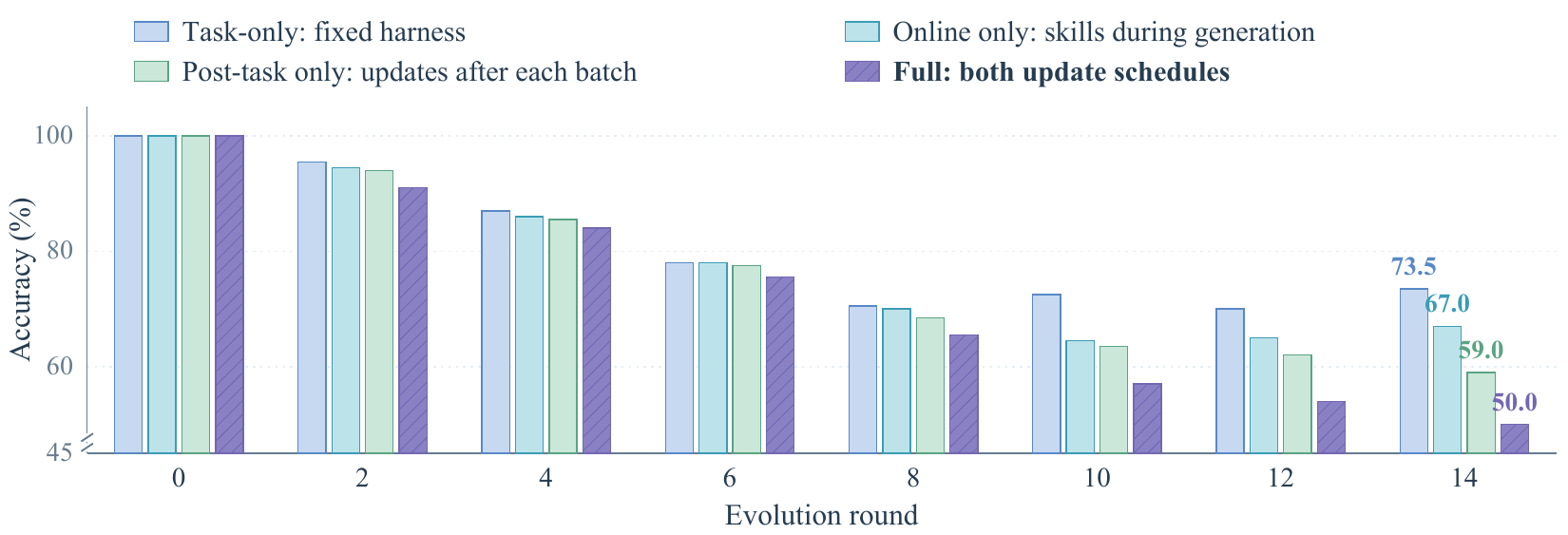}
    \caption{\textbf{Update ablations.} Four configurations on 50 shared seeds:
    fixed harness, online skill updates, post-task updates with candidate
    evaluation, and both schedules. Lower DeepSeek-V4-Pro accuracy
    indicates harder tasks; the y-axis starts at 45\%.}
    \label{fig:update-ablations}
\end{figure*}

Online only reaches 67.0\% at round~14, versus 73.5\% for Task-only,
but regresses from 64.5\% at round~10. The curve does not identify
skill quality as the cause. Post-task only reaches 59.0\%, while Full
reaches 50.0\%: improvements of 9.0 points over Post-task only and
23.5 points over Task-only (Figure~\ref{fig:update-ablations}). These
results support complementary update schedules, without isolating the
contribution of candidate screening within post-task optimization.

Full also improves over Online only by 17.0 points at round~14.
Together, the two comparisons support complementary roles: online
adaptation makes failure-derived skills available within the current
batch, while post-task optimization can reconsider the accumulated
library and the workflow that uses it. The combined configuration
benefits from both opportunities. These gains do not imply monotonic
improvement under either schedule or benefits from every new skill.

Late-round behavior further distinguishes the schedules. From round~10
to round~14, Full reduces solver accuracy from 57.0\% to 50.0\%,
while Post-task only improves from 63.5\% to 59.0\%. Both continue
making tasks harder as Online only regresses. This pattern is consistent
with a benefit from revisiting the accumulated harness, although the
finite trajectory does not establish eventual convergence for any
configuration.

\subsection{Harness Transfer and Generation Cost}
\label{sec:harness-evaluation}

\begin{figure*}[!t]
    \centering
    \includegraphics[width=\linewidth]{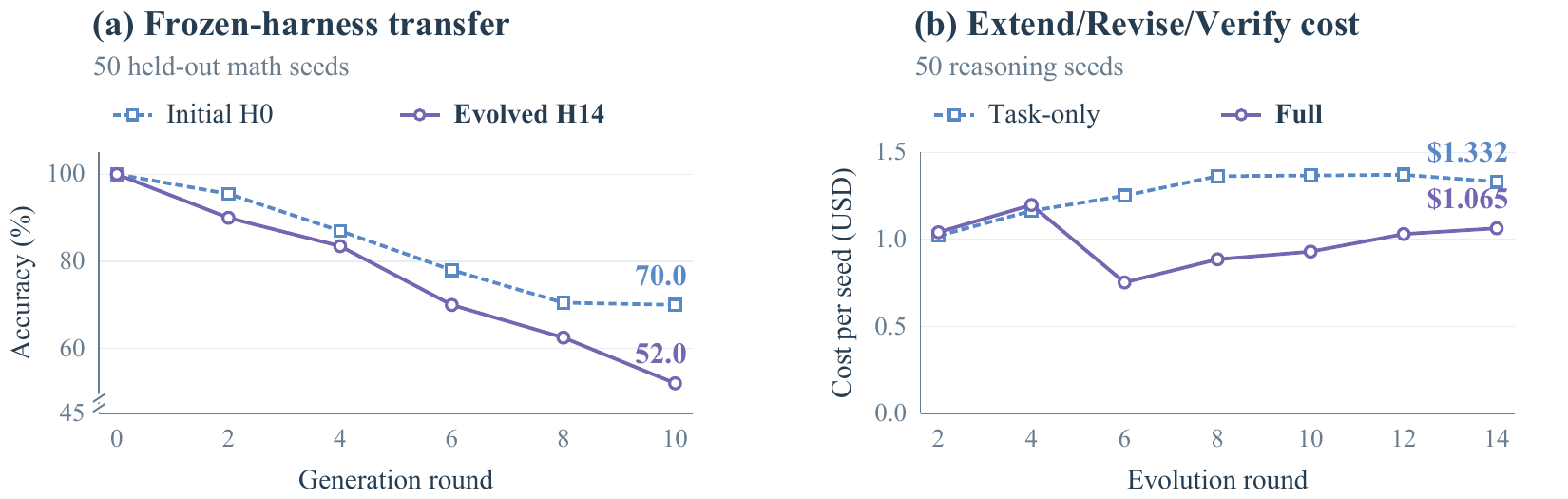}
    \caption{\textbf{Harness transfer and operator cost.}
    (a) Frozen harnesses on 50 shared held-out mathematics seeds.
    (b) Per-round Extend/Revise/Verify costs divided by 50 reasoning seeds;
    other harness and search costs are excluded. Full values are in
    Appendix~\ref{sec:harness-cost-records}.}
    \label{fig:harness-transfer-cost}
\end{figure*}

\paragraph{Frozen-harness transfer.}
Both snapshots, including their skill libraries, remain frozen on
50 mathematics seeds not used to evolve or select $H_{14}$. Subscripts
identify harness versions, not evaluation rounds. After ten recursive
generation rounds, $H_{14}$ yields 52.0\% solver accuracy versus 70.0\%
for $H_0$ (Figure~\ref{fig:harness-transfer-cost}a), supporting transfer
to these held-out seeds without further harness updates.
Because both harnesses restart from the same seeds, the 18.0-point
gap cannot be attributed to starting $H_{14}$ from its own harder
terminal tasks. Freezing the snapshots also prevents new online or
post-task adaptations during this comparison. The result supports
reuse of the evolved generation procedure beyond its original chains,
while evaluating transfer within mathematics under the same solver
rather than across solvers or domains.

\paragraph{Generation-operator cost.}
We divide each round's Extend/Revise/Verify fees by 50 reasoning seeds;
costs are not cumulative. Full costs slightly more at rounds 2 and 4,
but less at every reported round from 6 onward. At round 14, accuracy
is 50.0\% versus 73.5\% for Task-only, while cost is \$1.065 versus
\$1.332 per seed, a 20.0\% reduction (Figure~\ref{fig:harness-transfer-cost}b).
Excluding Director, skill-update, and search overhead, this measure is
narrower than $\bar C_{\mathrm{gen}}$ in Eq.~\eqref{eq:evaluation-metrics}
and cannot show total savings.

\subsection{Downstream Training Utility}
\label{sec:downstream}

\paragraph{GRPO on a 4B student.}
We train Qwen3.5-4B Thinking on 10K examples per data configuration.
Our data yields the highest accuracy on all four benchmarks
(Table~\ref{tab:grpo-4b}). The unweighted average reaches 62.7\%, versus
58.4\% for Qwen3.5-4B Thinking and 59.5\% for IQC, the strongest
data comparison by overall accuracy.
The gains over the checkpoint are 4.0, 4.8, 3.4, and 5.1 points on
APEX Shortlist, HMMT, LCB, and SciBench, respectively. Thus the overall
improvement accompanies gains in every evaluated domain. The strongest
alternative differs by benchmark: OpenMathInstruct-2 leads the data
comparisons on HMMT, OpenCodeReasoning on LCB, and IQC on APEX
Shortlist and SciBench. Our data exceeds each of these scores.

\paragraph{SFT on a 35B student.}
Our data leads on all four benchmarks
(Table~\ref{tab:sft-35b}); the unweighted average is 55.7\%, exceeding
both the checkpoint and MMIQC/IQC by 4.6 points using the displayed scores.
Together with GRPO, this supports utility under both training procedures.
OpenMathInstruct-2 and MMIQC/IQC improve APEX Shortlist and HMMT
but reduce LCB accuracy, whereas OpenCodeReasoning improves LCB but
reduces the three mathematics scores. Our configuration improves all
four scores, with gains of 3.1, 5.4, 5.8, and 4.3 points, respectively.
The aggregate gain therefore reflects improvements across both mathematics
and coding. These comparisons characterize the evaluated data mixtures
and training settings, rather than a universal ranking of source datasets.

\paragraph{APEX performance with a 27B student.}
Qwen3.8-27B improves from 56.3\% to
62.5\% mean-16 accuracy on APEX 2025, a 6.2-point gain using the displayed scores
(Figure~\ref{fig:intro-results}b; Appendix~\ref{sec:apex-reference}).
We evaluate all compared models with MathArena
\citep{dekoninck2026matharena}, using 16 attempts per problem.
Comparisons do not imply equal training or inference costs.

\paragraph{Scope of the evidence.}
Isolating the downstream contribution of harness evolution requires
matched-token training on our fixed-harness and evolved-harness data.
Total evolution cost, candidate validity, fallback rates, and
independent-run uncertainty remain unreported. Transfer to held-out
seeds and lower operator costs do not establish transfer to independent
solvers or end-to-end compute efficiency.

\section{Conclusion}
\label{sec:conclusion}

We introduced task--harness co-evolution for progressively harder
reasoning-data synthesis. The framework combines Online self-improvement
from intermediate solver failures with Post-task self-improvement
of skills, prompts, and workflows. Candidate evaluation and reversible snapshots
govern batch-level adoption, while verifier logic remains fixed.
Across mathematics, coding, and science, fixed-solver accuracy falls
from 100.0\% to 54.8\% by round~14. The update ablation further
shows a 23.5-point accuracy reduction relative to fixed-harness
recursion at round~14, supporting the combination of online and
batch-level adaptation. Downstream SFT and GRPO results establish
training utility across the evaluated benchmarks; a 27B student
fine-tuned on 10K mathematics examples reaches 62.5\% mean-16 accuracy
on APEX. These findings identify the synthesis harness as a useful
target of recursive improvement. Frozen evaluation on shared held-out
seeds further shows that the evolved harness can generate harder tasks
without additional adaptation. This result links the difficulty gains
to a reusable generation procedure, beyond the particular task chains
used to evolve it.

Future work should compare fixed- and evolved-harness data under matched
training tokens, and measure total generation cost, candidate validity,
and difficulty transfer to independent solvers.

\section*{AI use statement}

Generative AI tools assisted with synthetic data generation, code
development, experiment-log organization and analysis, manuscript
drafting, and language editing.
The authors are responsible for independently verifying all algorithmic
claims, references, experimental results, and final manuscript content.

\clearpage
\bibliographystyle{plainnat}
\bibliography{references,experiments/references}

@misc{matharena2025apex,
  author = {Dekoninck, Jasper and Jovanovi{\'c}, Nikola and Petrov, Ivo and Vechev, Martin},
  title = {{MathArena Apex}: Unconquered Final-Answer Problems},
  year = {2025},
  url = {https://matharena.ai/apex/},
  note = {Methodology: 16 attempts per problem for directly prompted models; average success over problems and attempts. Accessed September 24, 2026}
}

@article{dekoninck2026matharena,
  title = {Beyond Benchmarks: {MathArena} as an Evaluation Platform for Mathematics with {LLMs}},
  author = {Dekoninck, Jasper and Jovanovi{\'c}, Nikola and Gehrunger, Tim and R{\"o}gnvaldsson, K{\'a}ri and Petrov, Ivo and Sun, Chenhao and Vechev, Martin},
  journal = {arXiv preprint arXiv:2605.00674},
  year = {2026},
  url = {https://arxiv.org/abs/2605.00674}
}

@article{ahmad2025opencodereasoning,
  title = {{OpenCodeReasoning}: Advancing Data Distillation for Competitive Coding},
  author = {Ahmad, Wasi Uddin and Narenthiran, Sean and Majumdar, Somshubra and Ficek, Aleksander and Jain, Siddhartha and Huang, Jocelyn and Noroozi, Vahid and Ginsburg, Boris},
  journal = {arXiv preprint arXiv:2504.01943},
  year = {2025},
  url = {https://arxiv.org/abs/2504.01943}
}

@article{liu2024iqc,
  title = {Augmenting Math Word Problems via Iterative Question Composing},
  author = {Liu, Haoxiong and Zhang, Yifan and Luo, Yifan and Yao, Andrew Chi-Chih},
  journal = {arXiv preprint arXiv:2401.09003},
  year = {2024},
  url = {https://arxiv.org/abs/2401.09003}
}

@article{liu2026wildsci,
  title = {{WildSci}: Advancing Scientific Reasoning from In-the-Wild Literature},
  author = {Liu, Tengxiao and Nathani, Deepak and Li, Zekun and Yang, Kevin and Wang, William Yang},
  journal = {arXiv preprint arXiv:2601.05567},
  year = {2026},
  url = {https://arxiv.org/abs/2601.05567}
}

@article{yu2023metamath,
  title = {{MetaMath}: Bootstrap Your Own Mathematical Questions for Large Language Models},
  author = {Yu, Longhui and Jiang, Weisen and Shi, Han and Yu, Jincheng and Liu, Zhengying and Zhang, Yu and Kwok, James T. and Li, Zhenguo and Weller, Adrian and Liu, Weiyang},
  journal = {arXiv preprint arXiv:2309.12284},
  year = {2023},
  url = {https://arxiv.org/abs/2309.12284}
}

@article{jain2024livecodebench,
  title = {{LiveCodeBench}: Holistic and Contamination Free Evaluation of Large Language Models for Code},
  author = {Jain, Naman and Han, King and Gu, Alex and Li, Wen-Ding and Yan, Fanjia and Zhang, Tianjun and Wang, Sida and Solar-Lezama, Armando and Sen, Koushik and Stoica, Ion},
  journal = {arXiv preprint arXiv:2403.07974},
  year = {2024},
  url = {https://arxiv.org/abs/2403.07974}
}

@article{wang2024scibench,
  title = {{SciBench}: Evaluating College-Level Scientific Problem-Solving Abilities of Large Language Models},
  author = {Wang, Xiaoxuan and Hu, Ziniu and Lu, Pan and Zhu, Yanqiao and Zhang, Jieyu and Subramaniam, Satyen and Loomba, Arjun R. and Zhang, Shichang and Sun, Yizhou and Wang, Wei},
  journal = {arXiv preprint arXiv:2307.10635},
  year = {2024},
  url = {https://arxiv.org/abs/2307.10635}
}

@inproceedings{rein2024gpqa,
  title     = {{GPQA}: A Graduate-Level Google-Proof {Q\&A} Benchmark},
  author    = {Rein, David and Hou, Betty Li and
               Stickland, Asa Cooper and Petty, Jackson and
               Pang, Richard Yuanzhe and Dirani, Julien and
               Michael, Julian and Bowman, Samuel R.},
  booktitle = {First Conference on Language Modeling},
  year      = {2024},
  url       = {https://openreview.net/forum?id=Ti67584b98}
}

@misc{phan2025humanity,
  title         = {Humanity's Last Exam},
  author        = {Phan, Long and Gatti, Alice and Han, Ziwen and others},
  year          = {2025},
  eprint        = {2501.14249},
  archivePrefix = {arXiv},
  primaryClass  = {cs.LG},
  url           = {https://arxiv.org/abs/2501.14249}
}

@misc{wang2026frontierscience,
  title         = {{FrontierScience}: Evaluating {AI}'s Ability to
                   Perform Expert-Level Scientific Tasks},
  author        = {Wang, Miles and Lin, Robi and Hu, Kat and
                   Jiao, Joy and Chowdhury, Neil and Chang, Ethan and
                   Patwardhan, Tejal},
  year          = {2026},
  eprint        = {2601.21165},
  archivePrefix = {arXiv},
  primaryClass  = {cs.AI},
  url           = {https://arxiv.org/abs/2601.21165}
}

@inproceedings{wang2023selfinstruct,
  title     = {Self-Instruct: Aligning Language Models with
               Self-Generated Instructions},
  author    = {Wang, Yizhong and Kordi, Yeganeh and Mishra, Swaroop and
               Liu, Alisa and Smith, Noah A. and Khashabi, Daniel and
               Hajishirzi, Hannaneh},
  booktitle = {Proceedings of the 61st Annual Meeting of the
               Association for Computational Linguistics
               (Volume 1: Long Papers)},
  year      = {2023},
  pages     = {13484--13508},
  publisher = {Association for Computational Linguistics},
  doi       = {10.18653/v1/2023.acl-long.754},
  url       = {https://aclanthology.org/2023.acl-long.754/}
}

@inproceedings{toshniwal2025openmathinstruct2,
  title     = {{OpenMathInstruct-2}: Accelerating {AI} for Math with
               Massive Open-Source Instruction Data},
  author    = {Toshniwal, Shubham and Du, Wei and Moshkov, Ivan and
               Kisacanin, Branislav and Ayrapetyan, Alexan and Gitman, Igor},
  booktitle = {The Thirteenth International Conference on Learning
               Representations},
  year      = {2025},
  url       = {https://proceedings.iclr.cc/paper_files/paper/2025/hash/302ce0673c00aee2cf84bb43d0117553-Abstract-Conference.html}
}

@misc{wu2026benchevolver,
  title         = {{BenchEvolver}: Frontier Task Synthesis via
                   Solution-Centric Evolution},
  author        = {Wu, Yangzhen and Li, Aaron J. and Ma, Wenjie and
                   Cao, Li and Zhou, Ziheng and Cemri, Mert and
                   Liu, Shu and Xiu, Yuran and Yan, Chenxiao and
                   Zhao, Haikun and Yu, Bin and Stoica, Ion and
                   Song, Dawn},
  year          = {2026},
  eprint        = {2606.01286},
  archivePrefix = {arXiv},
  primaryClass  = {cs.SE},
  url           = {https://arxiv.org/abs/2606.01286}
}

@misc{li2026rst,
  title         = {Recursive Synthesis for Long-Horizon Terminal Tasks},
  author        = {Li, Zhongzhi and Shi, Yucheng and Li, Zongxia and
                   Wang, Ruhan and Li, Anhao and Huang, Zixun and
                   Yang, Junyao and Ke, Lei and Liu, Ninghao and
                   Mi, Haitao and Liang, Leowei},
  year          = {2026},
  eprint        = {2608.05466},
  archivePrefix = {arXiv},
  primaryClass  = {cs.AI},
  url           = {https://arxiv.org/abs/2608.05466}
}

@misc{kulikov2026autodata,
  title         = {{Autodata}: An agentic data scientist to create
                   high quality synthetic data},
  author        = {Kulikov, Ilia and Whitehouse, Chenxi and Wu, Tianhao and
                   Nie, Yixin and Saha, Swarnadeep and Helenowski, Eryk and
                   Yuan, Weizhe and Golovneva, Olga and Lanchantin, Jack and
                   Bachrach, Yoram and Foerster, Jakob and Li, Xian and
                   Fang, Han and Sukhbaatar, Sainbayar and Weston, Jason},
  year          = {2026},
  eprint        = {2606.25996},
  archivePrefix = {arXiv},
  primaryClass  = {cs.AI},
  url           = {https://arxiv.org/abs/2606.25996}
}

@misc{yang2026datafoundry,
  title         = {{DataFoundry}: Evolving Data Preparators via
                   Recursive Self-Improvement},
  author        = {Yang, Cehao and Wu, Xiaojun and Lin, Xueyuan and
                   Xu, Chengjin and Jiang, Xuhui and Xiong, Hui and
                   Guo, Jian},
  year          = {2026},
  eprint        = {2608.29966},
  archivePrefix = {arXiv},
  primaryClass  = {cs.CL},
  url           = {https://arxiv.org/abs/2608.29966}
}

@misc{lee2026rhi,
  title         = {Recursive Harness Self-Improvement},
  author        = {Lee, Hyunin and Xu, Jinglue and Seely, Jeffrey and
                   Lee, Donghyun and Zaharia, Matei and Tang, Yujin},
  year          = {2026},
  eprint        = {2607.15524},
  archivePrefix = {arXiv},
  primaryClass  = {cs.LG},
  url           = {https://arxiv.org/abs/2607.15524}
}

@misc{zhang2026hyperagents,
  title         = {Hyperagents},
  author        = {Zhang, Jenny and Zhao, Bingchen and Yang, Wannan and
                   Foerster, Jakob and Clune, Jeff and Jiang, Minqi and
                   Devlin, Sam and Shavrina, Tatiana},
  year          = {2026},
  eprint        = {2603.19461},
  archivePrefix = {arXiv},
  primaryClass  = {cs.AI},
  url           = {https://arxiv.org/abs/2603.19461}
}

@inproceedings{xu2024wizardlm,
  title     = {{WizardLM}: Empowering Large Pre-Trained Language Models
               to Follow Complex Instructions},
  author    = {Xu, Can and Sun, Qingfeng and Zheng, Kai and Geng, Xiubo
               and Zhao, Pu and Feng, Jiazhan and Tao, Chongyang
               and Lin, Qingwei and Jiang, Daxin},
  booktitle = {International Conference on Learning Representations},
  year      = {2024},
  url       = {https://arxiv.org/abs/2304.12244}
}

@inproceedings{zeng2024autoevol,
  title     = {Automatic Instruction Evolving for Large Language Models},
  author    = {Zeng, Weihao and Xu, Can and Zhao, Yingxiu and Lou,
               Jian-Guang and Chen, Weizhu},
  booktitle = {Proceedings of the 2024 Conference on Empirical Methods in
               Natural Language Processing},
  pages     = {6998--7018},
  year      = {2024},
  doi       = {10.18653/v1/2024.emnlp-main.397},
  url       = {https://aclanthology.org/2024.emnlp-main.397/}
}

@article{zhao2025absolutezero,
  title   = {Absolute Zero: Reinforced Self-Play Reasoning with Zero Data},
  author  = {Zhao, Andrew and Wu, Yiran and Yue, Yang and Wu, Tong
             and Xu, Quentin and Lin, Matthieu and Wang, Shenzhi
             and Wu, Qingyun and Zheng, Zilong and Huang, Gao},
  journal = {arXiv preprint arXiv:2505.03335},
  year    = {2025},
  url     = {https://arxiv.org/abs/2505.03335}
}

@inproceedings{hu2024adas,
  title     = {Automated Design of Agentic Systems},
  author    = {Hu, Shengran and Lu, Cong and Clune, Jeff},
  booktitle = {International Conference on Learning Representations},
  year      = {2025},
  url       = {https://proceedings.iclr.cc/paper_files/paper/2025/hash/36b7acf6f6010652b3f2a433774a66fe-Abstract-Conference.html}
}

@inproceedings{zhang2024aflow,
  title   = {{AFlow}: Automating Agentic Workflow Generation},
  author  = {Zhang, Jiayi and Xiang, Jinyu and Yu, Zhaoyang
             and Teng, Fengwei and Chen, Xionghui and Chen, Jiaqi
             and Zhuge, Mingchen and Cheng, Xin and Hong, Sirui
             and Wang, Jinlin and Zheng, Bingnan and Liu, Bang
             and Luo, Yuyu and Wu, Chenglin},
  booktitle = {International Conference on Learning Representations},
  year    = {2025},
  url     = {https://proceedings.iclr.cc/paper_files/paper/2025/hash/5492ecbce4439401798dcd2c90be94cd-Abstract-Conference.html}
}

@article{lee2026metaharness,
  title   = {Meta-Harness: End-to-End Optimization of Model Harnesses},
  author  = {Lee, Yoonho and Nair, Roshen and Zhang, Qizheng
             and Lee, Kangwook and Khattab, Omar and Finn, Chelsea},
  journal = {arXiv preprint arXiv:2603.28052},
  year    = {2026},
  url     = {https://arxiv.org/abs/2603.28052}
}

@inproceedings{hendrycks2021math,
  title     = {Measuring Mathematical Problem Solving With the {MATH} Dataset},
  author    = {Hendrycks, Dan and Burns, Collin and Kadavath, Saurav and
               Arora, Akul and Basart, Steven and Tang, Eric and Song, Dawn
               and Steinhardt, Jacob},
  booktitle = {Advances in Neural Information Processing Systems},
  year      = {2021},
  url       = {https://arxiv.org/abs/2103.03874}
}

@misc{glazer2024frontiermath,
  title         = {FrontierMath: A Benchmark for Evaluating Advanced Mathematical Reasoning in AI},
  author        = {Glazer, Elliot and Erdil, Ege and Besiroglu, Tamay and
                   Chicharro, Diego and Chen, Evan and Gunning, Alex and
                   Olsson, Caroline Falkman and Denain, Jean-Stanislas and
                   Ho, Anson and Santos, Emily de Oliveira and others},
  year          = {2024},
  eprint        = {2411.04872},
  archivePrefix = {arXiv},
  primaryClass  = {cs.AI},
  url           = {https://arxiv.org/abs/2411.04872}
}

@misc{tian2024scicode,
  title         = {SciCode: A Research Coding Benchmark Curated by Scientists},
  author        = {Tian, Minyang and Gao, Luyu and Zhang, Shizhuo Dylan and
                   Chen, Xinan and Fan, Cunwei and Guo, Xuefei and Haas,
                   Roland and Ji, Pan and others},
  year          = {2024},
  eprint        = {2407.13168},
  archivePrefix = {arXiv},
  primaryClass  = {cs.AI},
  url           = {https://arxiv.org/abs/2407.13168}
}

@article{trinh2024alphageometry,
  title   = {Solving Olympiad Geometry without Human Demonstrations},
  author  = {Trinh, Trieu H. and Wu, Yuhuai and Le, Quoc V. and He, He and
             Luong, Thang},
  journal = {Nature},
  volume  = {625},
  pages   = {476--482},
  year    = {2024},
  doi     = {10.1038/s41586-023-06747-5},
  url           = {https://doi.org/10.1038/s41586-023-06747-5}
}

@article{hubert2026alphaproof,
  title   = {Olympiad-Level Formal Mathematical Reasoning with Reinforcement Learning},
  author  = {Hubert, Thomas and Mehta, Rishi and Sartran, Laurent and
             Horv{\'a}th, Mikl{\'o}s Z. and others},
  journal = {Nature},
  volume  = {651},
  pages   = {607--613},
  year    = {2026},
  doi     = {10.1038/s41586-025-09833-y},
  url           = {https://doi.org/10.1038/s41586-025-09833-y}
}

@misc{guo2025deepseekr1,
  title         = {DeepSeek-R1: Incentivizing Reasoning Capability in LLMs via Reinforcement Learning},
  author        = {{DeepSeek-AI}},
  year          = {2025},
  eprint        = {2501.12948},
  archivePrefix = {arXiv},
  primaryClass  = {cs.CL},
  url           = {https://arxiv.org/abs/2501.12948}
}

@misc{novikov2025alphaevolve,
  title         = {AlphaEvolve: A Coding Agent for Scientific and Algorithmic Discovery},
  author        = {Novikov, Alexander and Vu, Ngân and Eisenberger, Marvin and
                   Dupont, Emilien and Huang, Po-Sen and Wagner, Adam Zsolt
                   and others},
  year          = {2025},
  eprint        = {2506.13131},
  archivePrefix = {arXiv},
  primaryClass  = {cs.AI},
  url           = {https://arxiv.org/abs/2506.13131}
}

@inproceedings{zelikman2022star,
  title     = {{STaR}: Bootstrapping Reasoning With Reasoning},
  author    = {Zelikman, Eric and Wu, Yuhuai and Mu, Jesse and Goodman, Noah D.},
  booktitle = {Advances in Neural Information Processing Systems},
  volume    = {35},
  pages     = {15476--15488},
  year      = {2022},
  url       = {https://proceedings.nips.cc/paper_files/paper/2022/hash/639a9a172c044fbb64175b5fad42e9a5-Abstract-Conference.html}
}

@article{singh2024restem,
  title   = {Beyond Human Data: Scaling Self-Training for Problem-Solving
             with Language Models},
  author  = {Singh, Avi and Co-Reyes, John D. and Agarwal, Rishabh and others},
  journal = {Transactions on Machine Learning Research},
  year    = {2024},
  url     = {https://arxiv.org/abs/2312.06585}
}

@misc{huang2025rzero,
  title         = {R-Zero: Self-Evolving Reasoning LLM from Zero Data},
  author        = {Huang, Chengsong and Yu, Wenhao and Wang, Xiaoyang and
                   Zhang, Hongming and Li, Zongxia and Li, Ruosen and
                   Huang, Jiaxin and Mi, Haitao and Yu, Dong},
  year          = {2025},
  eprint        = {2508.05004},
  archivePrefix = {arXiv},
  primaryClass  = {cs.AI},
  url           = {https://arxiv.org/abs/2508.05004}
}

@inproceedings{fernando2023promptbreeder,
  title         = {Promptbreeder: Self-Referential Self-Improvement via
                   Prompt Evolution},
  author        = {Fernando, Chrisantha and Banarse, Dylan Sunil and Michalewski,
                   Henryk and Osindero, Simon and Rockt\"aschel, Tim},
  booktitle     = {Proceedings of the 41st International Conference on Machine Learning},
  pages         = {13481--13544},
  year          = {2024},
  url           = {https://proceedings.mlr.press/v235/fernando24a.html}
}

@inproceedings{zelikman2023stop,
  title     = {Self-Taught Optimizer ({STOP}): Recursively Self-Improving
               Code Generation},
  author    = {Zelikman, Eric and Lorch, Eliana and Mackey, Lester and
               Kalai, Adam Tauman},
  booktitle = {First Conference on Language Modeling},
  year      = {2024},
  url       = {https://arxiv.org/abs/2310.02304}
}

@misc{zhang2025dgm,
  title         = {Darwin G\"odel Machine: Open-Ended Evolution of
                   Self-Improving Agents},
  author        = {Zhang, Jenny and Hu, Shengran and Lu, Cong and Lange,
                   Robert and Clune, Jeff},
  year          = {2025},
  eprint        = {2505.22954},
  archivePrefix = {arXiv},
  primaryClass  = {cs.AI},
  url           = {https://arxiv.org/abs/2505.22954}
}

\clearpage
\appendix
\section{Complete Evolution Procedure}
\label{sec:method-algorithm}

Algorithm~\ref{alg:evolution} integrates Online self-improvement and Post-task self-improvement.
The latter includes post-batch skill
updates, joint revision, candidate evaluation, and bounded retries.
The online snapshot is the default next-round harness; only an accepted
candidate replaces it. Thus three rejections discard all batch-level
changes without removing skills acquired during generation.

\begin{algorithm}[H]
\caption{Recursive Task Synthesis with an Adaptive Harness}
\label{alg:evolution}
\small
\begin{algorithmic}[1]
\Require Sources $\mathcal C$, candidate seeds $\mathcal Q$, benchmark registry $\mathcal K$;
\Statex \hspace{\algorithmicindent} verifier $V$, constraints $\Gamma$, goal $g$, generation/evaluation budgets, $N_b$ rounds.
\State $(\mathcal C_0,\mathcal T)\gets\Call{Curate}{\mathcal C,\mathcal Q;\mathcal K}$ \Comment{Exclude held-out families before distillation}
\State $H\gets\Call{Initialize}{\mathcal C_0}$; $\mathcal B\gets\{\Call{Snapshot}{H,\mathcal T}\}$ \Comment{Screened skills and seeds}
\For{$n=1,\ldots,N_b$}
    \State $H^{\mathrm{parent}}\gets H$; $H^{\mathrm w}\gets\Call{Copy}{H}$; $x\gets\Call{StartBatch}{\mathcal T}$
    \While{unfinished tasks and generation budget remain} \Comment{Online self-improvement}
        \State $(x,e)\gets\Call{DirectorStep}{x,H^{\mathrm w},V}$; $\Call{Log}{e}$
        \If{$e$ contains an attributed solver-miss trace}
            \State $H^{\mathrm w}.\mathcal L\gets\Call{Admit}{H^{\mathrm w}.\mathcal L,\operatorname{MetaSkill}(e;\Gamma_s)}$
        \EndIf \Comment{Next Director call reads the new library}
    \EndWhile
    \State $(\widetilde{\mathcal T},\mathcal Z,\mathcal M)\gets\Call{FinalizeBatch}{x;V}$
    \State $\mathcal T^+\gets\Call{AdvanceOrRetainValidSeeds}{\mathcal T,\widetilde{\mathcal T}}$
    \State $H^{\mathrm{online}}\gets\Call{Snapshot}{H^{\mathrm w}}$; $H\gets H^{\mathrm{online}}$; $a\gets\texttt{online-only}$
    \State $(\mathcal E,\mathcal F)\gets\Call{InLoopTest}{\mathcal T^+}$ \Comment{Post-task self-improvement}
    \State $\widehat H\gets\Call{Copy}{H^{\mathrm{online}}}$
    \State $\widehat H.\mathcal L\gets\Call{Admit}{\widehat H.\mathcal L,\operatorname{MetaSkill}(\mathcal F;\Gamma_s)}$
    \State $S\gets\Call{FreezeGateCohort}{50\text{ per domain}}$; $e_o\gets\Call{EvaluateOneRound}{H^{\mathrm{online}},S}$; $\mathcal R\gets\varnothing$
    \Statex \hspace{2\algorithmicindent} \Comment{Gate: 150 seeds; four solver attempts/task}
    \For{$j=1,2,3$} \Comment{Initial candidate and at most two retries, same $S$}
        \If{evaluation budget exhausted or baseline evaluation incomplete}
            \State \textbf{break} \Comment{Retain the online snapshot}
        \EndIf
        \State $H'\gets\Call{MetaHarness}{\operatorname{Copy}(\widehat H),\mathcal Z,\mathcal M,\mathcal E;\Gamma,g,\mathcal R}$
        \State $e'\gets\Call{CheckedEvaluateOneRound}{H',S}$ \Comment{Isolated probe, same limits}
        \If{$e'$ is complete and $\Call{Accept}{e',e_o;\Gamma,g}$ satisfies Eq.~\eqref{eq:acceptance}}
            \State $H\gets H'$; $a\gets\texttt{accepted}$; \textbf{break}
        \EndIf
        \State $\Call{Append}{\mathcal R,\operatorname{RejectReport}(e',e_o)}$
    \EndFor \Comment{After three rejections, $H$ remains $H^{\mathrm{online}}$}
    \State $\mathcal B\gets\Call{Archive}{\mathcal B,H^{\mathrm{parent}},H,\mathcal T^+,a,\mathcal R,\text{evaluations and costs}}$
    \State $\mathcal T\gets\mathcal T^+$ \Comment{Do not advance using probe outputs}
\EndFor
\State $\Call{AuditTrainingExport}{\mathcal T,H.\mathcal L;\mathcal K}$ \Comment{Retain the evaluation cohort}
\State \Return $(H,\mathcal T,\mathcal B)$
\end{algorithmic}
\end{algorithm}

\clearpage
\section{Method Protocol Details}
\label{sec:method-protocol}

\paragraph{Curation and overlap control.}
We freeze benchmark versions before screening both candidate seeds and
skill-distillation sources. Normalized exact duplicates and confirmed
benchmark overlaps are removed; lexical and semantic matching flags
near-duplicates and paraphrases for review. Shared concepts alone do
not constitute overlap. Duplicate seed families remain within a single
split when forming evolution and held-out pools. Only retained
materials enter the generation and acquisition agents; benchmark items
and solutions remain confined to the isolated overlap checker. Source
identifiers and filtering decisions are retained for auditing.
Distilled skill examples and final training exports undergo further
overlap checks. Training-export deduplication does not change the fixed
seed cohort used for difficulty measurement. These rules specify the
curation protocol; they do not establish the absence of contamination
without the corresponding audit records.

\paragraph{Skill representation and attribution.}
Each skill describes a reusable question-construction transformation.
The field \texttt{intent} identifies the reasoning capability to probe;
\texttt{method} describes the transformation;
\texttt{difficulty\_effect} states its intended effect on difficulty;
and \texttt{tool\_hint} identifies supporting tools or procedures.
Skill admission checks applicability, reusability, and supporting
evidence. A solver miss must be distinguished from an invalid problem,
an incorrect reference answer, or a tool failure, which instead calls
for repair. Trace and skill identifiers preserve provenance and prevent
duplicate processing. The operators $\operatorname{Meta}_s$ and
$\operatorname{Meta}_h$ denote skill-update and harness-revision roles;
they need not use different underlying models.

\paragraph{Isolated candidate evaluation.}
After the batch, freeze the shared gate cohort $S_n$: 50 current retained
tasks each from mathematics, coding, and science (150 in total).
This gate cohort is specified separately from the 50-seed cohorts used
to report the update ablation, frozen-harness transfer, and operator cost.
The online baseline and each candidate synthesize one additional round
from this set under matched model/tool settings and resource ceilings.
Inputs and once-computed baseline results remain fixed through the
initial candidate and at most two retries (three candidate evaluations).
Each probe copies the relevant snapshot, retains its full skill library,
and follows the same online-adaptation protocol. Probe-time skills are
neither shared nor written back; probe outputs do not advance production.
Equal libraries are needed only to isolate workflow edits in an ablation.
This current-task selection is not a held-out generalization test.
Each proposal starts from a fresh copy of $\widehat H$ plus rejection
reports, without accumulating rejected edits.

\paragraph{Acceptance metrics and budget.}
The gate uses 150 shared seed chains (50 per domain) and four solver attempts per
retained output, or 600 outcomes per harness. Mean accuracy gives each
seed equal weight. The 600 outcomes are clustered within 150 tasks,
not independent tasks. Relative difficulty is the ratio of the two cohort
accuracies; a decrease from 0.60 to 0.50 gives $r_D=1.2$.
Equation~\eqref{eq:acceptance} requires strictly lower accuracy and no
decrease in new-candidate validity. The cost-growth coefficient and
relative cap are both 1.5, fixed before search. In this example, a
candidate may cost up to 1.30 times the baseline. These are engineering
preferences for bounded spending, not statistically calibrated thresholds
or a claim of optimal cost efficiency.

If $\bar A(H_o)=0$, no candidate can achieve strictly lower accuracy.
If $\bar A(H')=0<\bar A(H_o)$, apply the 1.5 cost cap directly without
dividing by zero. More generally, after checking the strict accuracy
condition, the cost test is equivalently
\[
\begin{aligned}
\bar C_{\mathrm{gen}}(H')&\le1.5\bar C_{\mathrm{gen}}(H_o),\\
\bar C_{\mathrm{gen}}(H')\bar A(H')&\le
\bar C_{\mathrm{gen}}(H_o)\bigl[1.5\bar A(H_o)-0.5\bar A(H')\bigr].
\end{aligned}
\]
Difficulty uses the latest valid completed task from every original
seed chain. In contrast, $Q$ measures new-candidate validity before
fallback; missing new candidates count as invalid. This prevents
retaining valid predecessors from masking failed synthesis.

Input tasks, solver settings, absolute generation/evaluation ceilings,
and baseline results stay fixed throughout a search. The relative cost
allowance does not raise these ceilings. Stop at the first accepted
candidate, after three rejected candidates, or when the evaluation
budget expires; without acceptance, retain the online snapshot.
With one baseline and three complete candidate probes, difficulty
scoring uses at most 2,400 solver attempts, excluding generation-time
verification and other calls. Incomplete probes cannot be accepted.

\paragraph{Cost accounting and reporting.}
The generation cost in Eq.~\eqref{eq:evaluation-metrics} includes Director,
construction, revision, verification, and online skill-update calls,
including unsuccessful attempts. Total evolution cost additionally
includes initialization, post-batch evaluation and updates, and all
candidate-search calls, including rejected proposals and baseline
evaluations. Divide by the original seed count rather than the number
of newly successful generations. Evaluation performed solely for final
reporting is accounted for separately and never feeds back into
updates. Cost comparisons should report difficulty and absolute cost
separately, with consistent pricing or token accounting, and include a
matched-total-budget comparison. Passing the gate does not establish
lower absolute cost, lower cost per unit of difficulty, or lower total
evolution cost. Absolute limits remain necessary because the relative
cap only compares a candidate with its current online baseline.

\paragraph{Version history and fallback.}
The archive $\mathcal B$ stores complete harness and task snapshots,
parent identifiers, acceptance status, evaluation records, rejection
reports, and costs. Accepted candidates create new accepted versions;
three rejected candidates produce an online-only version retaining
$H^{\mathrm{online}}$. The latter does not represent a Critic-accepted
batch-level improvement. Both outcomes use the retained tasks as seeds
for the next round and preserve reproducible rollback points.

\clearpage
\section{Additional APEX Training and Reference Results}
\label{sec:apex-reference}

\paragraph{Before and after SFT.}
The Qwen3.8-27B starting checkpoint achieves 56.3\% on APEX, while
SFT on 10K synthesized mathematics examples reaches 62.5\%
(Table~\ref{tab:apex-27b-sft-gain}). The improvement is 6.2 percentage
points, computed from the displayed scores. Both results use mean-16
accuracy. This within-checkpoint comparison complements the 35B
training-data comparison in Table~\ref{tab:sft-35b}; it measures the
observed training gain, rather than isolating the contribution of
harness evolution to that gain.

\begin{table}[!tb]
    \centering
    \caption{\textbf{27B training gain on APEX.} Mean-16 accuracy (\%).
    Scores are rounded to one decimal; gains are computed from
    the displayed scores.}
    \label{tab:apex-27b-sft-gain}
    \small
    \setlength{\tabcolsep}{10pt}
    \begin{tabular}{lrr}
        \toprule
        \textbf{Qwen3.8-27B configuration} & \textbf{APEX} & \textbf{Gain (pp)} \\
        \midrule
        Before SFT & 56.3 & --- \\
        \textbf{SFT on our 10K math examples} & \textbf{62.5} & \textbf{+6.2} \\
        \bottomrule
    \end{tabular}
\end{table}

\paragraph{Selected reference models.}
Table~\ref{tab:apex-27b} compares our 27B SFT model with selected
reference models. We evaluate every listed model ourselves on APEX 2025
using the MathArena framework~\citep{dekoninck2026matharena}.
Our model uses 10K mathematics examples for SFT. Each reported score
averages 16 attempts per problem (mean-16), following the APEX
protocol~\citep{matharena2025apex}, rather than measuring pass@16.
Scores are displayed to one decimal, with gaps calculated
from the displayed scores. The 3.1-point gap to Kimi K3 (Think) describes point
estimates, not statistical equivalence. The ordering concerns only the
listed models, and equal evaluation counts do not imply equal compute.

\begin{table}[!tb]
    \centering
    \caption{\textbf{APEX reference comparison.} Mean-16 accuracy (\%)
    on APEX 2025, rounded to one decimal. All scores are from our own
    evaluations using MathArena. Only our 27B model receives additional
    training in this comparison.}
    \label{tab:apex-27b}
    \small
    \setlength{\tabcolsep}{12pt}
    \begin{tabular}{lr}
        \toprule
        \textbf{Model} & \textbf{APEX} \\
        \midrule
        Kimi K3 (Think) & 65.6 \\
        \textcolor{paperPurpleDark}{\textbf{Ours: Qwen3.8-27B + SFT}} & \textcolor{paperPurpleDark}{\textbf{62.5}} \\
        DeepSeek-v4-pro & 61.5 \\
        Gemini 3.1 Pro Preview & 60.9 \\
        GPT-5.4 (xhigh) & 54.2 \\
        DeepSeek-v4-flash-0731 & 44.8 \\
        Qwen3.7-max & 43.8 \\
        Claude-Opus-4.7 (xhigh) & 40.6 \\
        Claude-Opus-4.6 (High) & 34.9 \\
        Gemini 3.5 Flash & 32.3 \\
        DeepSeek-v4-Pro (Max) & 28.1 \\
        DeepSeek-v4-Flash (Max) & 27.1 \\
        \bottomrule
    \end{tabular}
\end{table}

\clearpage
\section{Numerical Difficulty and Ablation Results}
\label{sec:evolution-records}

Tables~\ref{tab:domain-progression} and~\ref{tab:task-harness-comparison}
retain all reported values underlying
Figures~\ref{fig:domain-progression} and~\ref{fig:update-ablations}.
The overall accuracy is recomputed from three equally weighted domains.
The four solver evaluations used in the domain experiment are not
independent evolution runs; no run-level uncertainty is inferred from
these aggregate values.

\begin{table*}[!t]
    \centering
    \caption{\textbf{Difficulty progression.} DeepSeek-V4-Pro accuracy
    (\%; lower is harder) over the fixed seed cohort. The overall row is
    recomputed from the three equally weighted domains.}
    \label{tab:domain-progression}
    \small
    \setlength{\tabcolsep}{3.5pt}
    \resizebox{\linewidth}{!}{%
    \begin{tabular}{lrrrrrrrrrrrr}
        \toprule
        \textbf{Domain} & $r{=}0$ & $1$ & $2$ & $3$ & $4$ & $5$ & $6$ & $7$ & $8$ & $9$ & $10$ & $14$ \\
        \midrule
        All (150) & 100.0 & 97.7 & 92.8 & 89.2 & 85.5 & 81.5 & 77.3 & 72.8 & 68.3 & 64.3 & 59.7 & 54.8 \\
        Math (50) & 100.0 & 98.5 & 91.0 & 87.5 & 84.0 & 80.0 & 75.5 & 70.0 & 65.5 & 62.0 & 57.0 & 50.0 \\
        Coding (50) & 100.0 & 97.0 & 93.5 & 90.0 & 86.0 & 82.5 & 78.0 & 74.5 & 70.0 & 66.5 & 62.0 & 58.5 \\
        Science (50) & 100.0 & 97.5 & 94.0 & 90.0 & 86.5 & 82.0 & 78.5 & 74.0 & 69.5 & 64.5 & 60.0 & 56.0 \\
        \bottomrule
    \end{tabular}}
\end{table*}

\begin{table*}[!t]
    \centering
    \caption{\textbf{Update ablations.} DeepSeek-V4-Pro accuracy
    (\%; lower is harder) on 50 shared seeds. Online only and Post-task only refer to the update mechanisms
    in Sections~\ref{sec:online} and~\ref{sec:batch}, respectively. Bold marks the lowest non-initial accuracy in
    each column.}
    \label{tab:task-harness-comparison}
    \small
    \setlength{\tabcolsep}{5pt}
    \resizebox{\linewidth}{!}{%
    \begin{tabular}{lrrrrrrrr}
        \toprule
        \textbf{Configuration} & $r{=}0$ & $2$ & $4$ & $6$ & $8$ & $10$ & $12$ & $14$ \\
        \midrule
        Task-only & 100.0 & 95.5 & 87.0 & 78.0 & 70.5 & 72.5 & 70.0 & 73.5 \\
        Online only & 100.0 & 94.5 & 86.0 & 78.0 & 70.0 & 64.5 & 65.0 & 67.0 \\
        Post-task only & 100.0 & 94.0 & 85.5 & 77.5 & 68.5 & 63.5 & 62.0 & 59.0 \\
        \textbf{Full} & 100.0 & \textbf{91.0} & \textbf{84.0} & \textbf{75.5} & \textbf{65.5} & \textbf{57.0} & \textbf{54.0} & \textbf{50.0} \\
        \bottomrule
    \end{tabular}}
\end{table*}

\clearpage
\section{Frozen-Harness and Cost Evaluation}
\label{sec:harness-cost-records}

\paragraph{Frozen-harness protocol.}
The two snapshots start from the same 50 mathematics seeds, which did
not participate in the evolution or selection of $H_{14}$. Skills,
prompts, workflows, and verification/revision placements remain frozen;
only the tasks evolve. Table~\ref{tab:frozen-harness} reports the fresh
generation trajectory, whose round index is distinct from the harness
version. These seeds are separate from the reasoning cohort used for
the operator-cost experiment. The $H_0$ round-10 accuracy of 70.0\%
belongs to this experiment and is not the 72.5\% task-only entry from
the update ablation.

\begin{table*}[!t]
    \centering
    \caption{\textbf{Frozen-harness comparison.} DeepSeek-V4-Pro
    accuracy (\%; lower is harder) on the tasks generated from 50 shared
    held-out mathematics seeds. Both harnesses remain frozen.}
    \label{tab:frozen-harness}
    \small
    \setlength{\tabcolsep}{9pt}
    \begin{tabular}{lrrrrrr}
        \toprule
        \textbf{Harness} & $r{=}0$ & $2$ & $4$ & $6$ & $8$ & $10$ \\
        \midrule
        Initial $H_0$ & 100.0 & 95.5 & 87.0 & 78.0 & 70.5 & 70.0 \\
        Evolved $H_{14}$ & 100.0 & \textbf{90.0} & \textbf{83.5} & \textbf{70.0} & \textbf{62.5} & \textbf{52.0} \\
        \bottomrule
    \end{tabular}
\end{table*}

\paragraph{Operator-cost scope.}
For each reported round $r$, the denominator is the original 50
reasoning seeds, rather than the number of newly successful tasks:
\[
\bar C_{\mathrm{ERV}}^{(r)}
=\frac{C_{\mathrm{Extend}}^{(r)}+C_{\mathrm{Revise}}^{(r)}
+C_{\mathrm{Verify}}^{(r)}}{50}.
\]
This measure includes only the reported generation-operator fees.
It excludes Director calls, online skill updates, post-task updates,
candidate search and evaluation (including rejected candidates),
initialization, and standalone reporting evaluations. It is therefore
not the full generation-cost numerator used for harness acceptance
or an estimate of total search cost. Values are for individual rounds;
unreported rounds are not interpolated or included in a cumulative sum.
No cost is assigned to the starting seed row.

\begin{table*}[!t]
    \centering
    \caption{\textbf{Accuracy and generation-operator cost.}
    DeepSeek-V4-Pro accuracy (\%) and per-round cost (USD per original
    seed) on 50 reasoning seeds. Lower is better for both measures.}
    \label{tab:operator-cost}
    \small
    \setlength{\tabcolsep}{5pt}
    \resizebox{\linewidth}{!}{%
    \begin{tabular}{llrrrrrrr}
        \toprule
        \textbf{Configuration} & \textbf{Measure} & $r{=}2$ & $4$ & $6$ & $8$ & $10$ & $12$ & $14$ \\
        \midrule
        Task-only & Accuracy & 95.5 & 87.0 & 78.0 & 70.5 & 72.5 & 70.0 & 73.5 \\
                  & Cost & 1.022 & 1.165 & 1.252 & 1.364 & 1.368 & 1.372 & 1.332 \\
        \addlinespace[3pt]
        Full & Accuracy & 91.0 & 84.0 & 75.5 & 65.5 & 57.0 & 54.0 & 50.0 \\
             & Cost & 1.042 & 1.199 & 0.754 & 0.887 & 0.931 & 1.032 & 1.065 \\
        \bottomrule
    \end{tabular}}
\end{table*}

At round 14, Full reduces solver accuracy by 23.5 percentage points
and operator cost by \$0.267 per seed (20.0\%). The cost difference
is computed from the reported three-decimal fees. The earlier
round-2 and round-4 operator costs are respectively 2.0\% and 2.9\%
higher for Full. These results do not quantify the cost of obtaining
$H_{14}$ or establish matched-total-budget superiority. Raw billing
records, pricing schedules, call counts, and independent-run uncertainty
are needed for a full cost audit.

\clearpage
\section{Experimental Setup}
\label{sec:experimental-setup}

\subsection{Models, Benchmarks, and Evaluation}

\paragraph{Synthesis models.}
Director, Extend, and Revise use DeepSeek-V4-Pro. Verify uses
Qwen3.8-Max, DeepSeek-V4-Pro, and GPT-5.6; the meta-agent uses
DeepSeek-V4-Pro and GPT-5.6. All listed synthesis and meta-agent models
run with Thinking enabled.

\paragraph{Evaluation and student models.}
We use DeepSeek-V4-Pro as the fixed solver for difficulty evaluation.
Students are Qwen3.5-4B Thinking for GRPO and Qwen3.5-35B-A3B and
Qwen3.8-27B for SFT.

\paragraph{Benchmarks.}
We evaluate on APEX/Shortlist, HMMT, LiveCodeBench v6
\citep{jain2024livecodebench}, and SciBench \citep{wang2024scibench}.
GRPO and SFT use HMMT February 2025 and 2026, respectively.
The evaluated LCB v6 and SciBench sets contain 454 and 695 problems,
respectively.

\paragraph{Evaluation protocols.}
We use the MathArena evaluation framework \citep{dekoninck2026matharena}.
All APEX scores, including those of the reference models, are obtained
in our own evaluations.
Difficulty averages four solver attempts per task; lower accuracy is
harder. All four GRPO benchmarks use mean-16 accuracy: the mean
correctness across 16 test attempts per problem, not pass@16.
Higher downstream accuracy is better; SFT protocols are specified
alongside the results.

\paragraph{Datasets and baselines.}
We compare with OpenMathInstruct-2
\citep{toshniwal2025openmathinstruct2}, OpenCodeReasoning
\citep{ahmad2025opencodereasoning}, and recursive IQC/MMIQC
\citep{liu2024iqc}; GRPO also includes WildSci \citep{liu2026wildsci}
and MetaMath \citep{yu2023metamath}.

\subsection{Training Configuration}

\paragraph{Controlled training comparisons.}
The 4B GRPO and both SFT experiments use \texttt{verl} on 16 NVIDIA
H20 GPUs, with shared hyperparameters across datasets within each
comparison. GRPO uses 10K training examples per data configuration.
The 35B comparison uses 12K examples per configuration
(ours: 10K mathematics and 2K coding); the 27B run uses 10K mathematics
examples. Matching hyperparameters and example counts does not imply
matched realized training tokens.

\begin{table*}[!t]
    \centering
    \caption{\textbf{Downstream training hyperparameters.} The SFT
    settings apply to both the 35B and 27B students. Sequence limits
    are token ceilings, not observed average lengths. Dashes denote
    values not specified here.}
    \label{tab:training-hyperparameters}
    \small
    \setlength{\tabcolsep}{7pt}
    \renewcommand{\arraystretch}{1.08}
    \begin{tabular}{@{}lll@{}}
        \toprule
        \textbf{Parameter} & \textbf{SFT (35B / 27B)} & \textbf{GRPO (4B)} \\
        \midrule
        Training hardware & $16\times$ NVIDIA H20 & $16\times$ NVIDIA H20 \\
        Epochs & 2 & 1 \\
        Learning rate & $10^{-5}$ & $10^{-6}$ \\
        Training batch size & 16 examples & 32 prompts \\
        Sequence limits & 65,536 total & \shortstack[l]{3,072 prompt;\\65,536 response} \\
        Distributed strategy & FSDP2 & FSDP \\
        Dynamic batching & Enabled & Enabled \\
        Gradient checkpointing & Enabled & Enabled \\
        Overlength handling & Error & Error \\
        \midrule
        \multicolumn{3}{@{}l}{\textbf{SFT optimizer and memory settings}} \\
        Warmup ratio & 0.05 & --- \\
        Learning-rate schedule & Cosine & --- \\
        Minimum LR / initial LR & 0.1 & --- \\
        Weight decay & 0.1 & --- \\
        Optimizer betas & $(0.9,\,0.95)$ & --- \\
        Gradient clipping & 1.0 & --- \\
        Model dtype & BF16 & --- \\
        Per-GPU micro-batch size & 1 & --- \\
        Per-GPU token ceiling & 65,536 & --- \\
        Sequence parallel size & 1 & --- \\
        \midrule
        \multicolumn{3}{@{}l}{\textbf{GRPO rollout and regularization settings}} \\
        Rollouts per prompt & --- & 8 \\
        Rollouts per full batch & --- & 256 \\
        PPO mini-batch size & --- & 32 \\
        Rollout backend & --- & vLLM \\
        Rollout temperature & --- & 1.0 \\
        Rollout tensor parallel size & --- & 1 \\
        vLLM GPU memory utilization & --- & 0.45 \\
        KL in reward & --- & Disabled \\
        KL loss coefficient & --- & 0.001 \\
        KL loss type & --- & \texttt{low\_var\_kl} \\
        Entropy coefficient & --- & 0 \\
        \bottomrule
    \end{tabular}
\end{table*}

\paragraph{Implementation.}
SFT enables padding removal, gradient checkpointing, activation
offloading, and resharding after the forward pass; compilation with
\texttt{torch.compile} is disabled. GRPO uses dynamic batching for
the actor and rollout log-probability computation. The actor keeps
parameters and optimizer states on device, while the reference model
offloads its parameters. KL regularization is applied through the
actor loss, separately from the reward.

\end{document}